\documentclass[10pt]{article} 
\usepackage[accepted]{tmlr}

\usepackage{amsmath,amsfonts,bm}

\def\eqref#1{(\ref{#1})}

\def\1{\bm{1}}

\DeclareMathAlphabet{\mathsfit}{\encodingdefault}{\sfdefault}{m}{sl}
\SetMathAlphabet{\mathsfit}{bold}{\encodingdefault}{\sfdefault}{bx}{n}

\usepackage[utf8]{inputenc} 
\usepackage[T1]{fontenc}    
\usepackage{hyperref}       
\usepackage{url}            
\usepackage{booktabs}       
\usepackage{multirow}
\usepackage{amsfonts}       
\usepackage{nicefrac}       
\usepackage{microtype}      
\usepackage{subcaption}
\usepackage{caption}
\usepackage{graphicx}
\usepackage{amsfonts, amsthm, mathtools}
\usepackage[ruled,vlined]{algorithm2e}
\usepackage{amsmath}
\usepackage{amssymb}
\usepackage{mathtools}
\usepackage{amsthm}
\usepackage{bm}
\usepackage{bbm}
\usepackage{comment}
\usepackage{makecell}

\usepackage{amssymb}
\usepackage{pifont}
\newcommand{\cmark}{\ding{51}}%
\newcommand{\xmark}{\ding{55}}%

\newtheorem{definition}{Definition}
\newtheorem{theorem}{Theorem}

\newtheorem{lemma}{Lemma}

\newtheorem{corollary}{Corollary}
\theoremstyle{definition}
\newtheorem{remark}{Remark}

\title{Adversarial Imitation Learning from Visual Observations \\using Latent Information}

\author{\name Vittorio Giammarino \email vgiammar@bu.edu \\
      \addr Division of Systems Engineering\\
      Boston University
      \AND
      \name James Queeney \email queeney@merl.com \\
      \addr Mitsubishi Electric Research Laboratories
      \AND
      \name Ioannis Ch. Paschalidis \email yannisp@bu.edu\\
      \addr Department of Electrical and Computer Engineering\\
      Division of Systems Engineering\\
      Faculty of Computing \& Data Sciences\\
      Boston University}

\def\month{05}  
\def\year{2024} 
\def\openreview{\url{https://openreview.net/forum?id=ydPHjgf6h0}} 

\begin{document}

\maketitle

\begin{abstract}
We focus on the problem of imitation learning from visual observations, where the learning agent has access to videos of experts as its sole learning source. The challenges of this framework include the absence of expert actions and the partial observability of the environment, as the ground-truth states can only be inferred from pixels. To tackle this problem, we first conduct a theoretical analysis of imitation learning in partially observable environments. We establish upper bounds on the suboptimality of the learning agent with respect to the divergence between the expert and the agent latent state-transition distributions. Motivated by this analysis, we introduce an algorithm called Latent Adversarial Imitation from Observations, which combines off-policy adversarial imitation techniques with a learned latent representation of the agent's state from sequences of observations. In experiments on high-dimensional continuous robotic tasks, we show that our model-free approach in latent space matches state-of-the-art performance. 
Additionally, we show how our method can be used to improve the efficiency of reinforcement learning from pixels by leveraging expert videos. To ensure reproducibility, we provide free access to all the learning curves and open-source our \href{https://github.com/VittorioGiammarino/AIL_from_visual_obs/tree/LAIfO}{code}. 
\end{abstract}

\section{Introduction}

Learning from videos represents a compelling opportunity for the future, as it offers a cost-effective and efficient way to teach autonomous agents new skills and behaviors. Compared to other methods, video recording is a faster and more flexible alternative for gathering data. Moreover, with the abundance of high-quality videos available on the internet, learning from videos has become increasingly accessible in recent years. However, despite the potential benefits, this approach remains challenging as it involves several technical problems that must be addressed simultaneously in order to succeed. These problems include representation learning, significant computational demands due to high-dimensional observation space, the partial observability of the decision process, and lack of expert actions. Our objective is to establish algorithms capable of overcoming all of these challenges, enabling the learning of complex robotics tasks directly from videos of experts.

Formally, our focus is on the problem of \emph{Visual Imitation from Observations (V-IfO)}. In V-IfO, the learning agent does not have access to a pre-specified reward function, and instead has to learn by imitating an expert's behavior. Additionally, in V-IfO, expert actions are not accessible during the learning process, and the pixel-based observations we obtain from video frames result in partial observability. The absence of expert actions and the partial observability of the environment distinguish V-IfO from other types of imitation from experts. Specifically, we identify three other frameworks previously addressed in the literature: \emph{Imitation Learning (IL)} \citep{atkeson1997robot, abbeel2004apprenticeship, ross2010efficient, reske2021imitation, giammarino2023combining} where states are fully observable and expert state-action pairs are accessible, \emph{Visual Imitation Learning (V-IL)} \citep{rafailov2021visual} which explores the idea of imitating directly from pixels but still assumes that expert actions are provided to the learning agent, and \emph{Imitation from Observations (IfO)} \citep{torabi2018generative, torabi2018behavioral} which retains full observability but considers only the availability of expert states. Table~\ref{table_1} summarizes these frameworks.

\begin{table}
\centering
\small
\caption{A summary of imitation from experts: Imitation Learning (IL), Imitation from Observations (IfO), Visual Imitation Learning (V-IL), and Visual Imitation from Observations (V-IfO).}
\label{table_1}
\begin{tabular}{l c c c c c c c c}\toprule
 & \multicolumn{2}{c}{IL} & \multicolumn{2}{c}{IfO} & \multicolumn{2}{c}{V-IL} & \multicolumn{2}{c}{V-IfO} \\
\cmidrule(lr){1-9}
Fully observable environment & \multicolumn{2}{c}{\cmark} & \multicolumn{2}{c}{\cmark} & \multicolumn{2}{c}{\xmark} & \multicolumn{2}{c}{\xmark} \\
Access to expert actions & \multicolumn{2}{c}{\cmark} & \multicolumn{2}{c}{\xmark} & \multicolumn{2}{c}{\cmark} & \multicolumn{2}{c}{\xmark} \\
\bottomrule
\end{tabular}
\end{table}

In order to address the V-IfO problem, this paper introduces both theoretical and algorithmic contributions. First, we provide a theoretical analysis of the problem and demonstrate that the suboptimality of the learning agent can be upper bounded by the divergence between the expert and the agent latent state-transition distributions. Our analysis motivates the reduction of the V-IfO problem to two subproblems: $(i)$~estimating a proper latent representation from sequences of observations and $(ii)$~efficiently minimizing the divergence between expert and agent distributions in this latent space. Next, we propose practical solutions to these subproblems. By doing so, we formalize a novel algorithm called \emph{Latent Adversarial Imitation from Observations (LAIfO)}, which tackles the divergence minimization step using off-policy adversarial imitation techniques \citep{ghasemipour2020divergence} and recovers a latent representation of the ground-truth state by means of observations stacking \citep{mnih2013playing, mnih2015human} and data augmentation \citep{laskin2020reinforcement, kostrikov2020image, yarats2021mastering}. We evaluate our algorithm on the DeepMind Control Suite \citep{tassa2018deepmind}, demonstrating that our model-free approach in latent space achieves state-of-the-art performance. 
We conclude by showing how LAIfO can be used on challenging environments, such as the humanoid from pixels \citep{tassa2018deepmind}, to improve Reinforcement Learning (RL) efficiency by leveraging expert videos.            

The remainder of the paper is organized as follows: Section~\ref{sec:related_work} provides a summary of the most related works to this paper. Section~\ref{sec:preliminaries} introduces notation and background on RL and IL. Section~\ref{sec:theoretical_analysis} provides a theoretical analysis of the V-IfO problem. Section~\ref{sec:LAIfO} introduces our algorithm, LAIfO, and outlines how it can leverage expert videos to improve data efficiency of RL from pixels. Finally, Section~\ref{sec:experiments} presents our experimental results and Section~\ref{sec:conclusion} concludes the paper providing a general discussion on our findings.

\section{Related work} 
\label{sec:related_work}
 
In recent years, several studies have focused on the IL problem \citep{atkeson1997robot, abbeel2004apprenticeship, ross2010efficient, reddy2019sqil, reske2021imitation, giammarino2023combining} and, in particular, on the generative adversarial IL framework \citep{ho2016generative} which has emerged as one of the most promising approaches for IL. Adversarial IL builds upon a vast body of work on inverse RL \citep{russell1998learning, ng2000algorithms, abbeel2004apprenticeship, syed2007game, ziebart2008maximum, syed2008apprenticeship}. The primary goal of inverse RL is to identify a reward function that enables expert trajectories (i.e., state-action pairs) to be optimal. The reward function obtained from the inverse RL step is then used to train agents in order to match the expert's expected reward. In the fully observable setting, adversarial IL was originally formalized in \citet{ho2016generative} and  \citet{fu2017learning}. It was later extended to the observation only setting in \citet{torabi2018generative} and to the visual setting in \cite{karnan2022adversarial}. Furthermore, the adversarial IfO problem has been theoretically analyzed in \citet{yang2019imitation} and \citet{cheng2021guaranteed}. Note that all of these studies are built upon on-policy RL \citep{schulman2017proximal}, which provides good learning stability but is known for poor sample efficiency. In recent works, this efficiency issue has been addressed by using off-policy RL algorithms in the adversarial optimization step \citep{haarnoja2018soft, lillicrap2015continuous}. These include DAC \citep{kostrikov2018discriminator}, SAM \citep{blonde2019sample}, and ValueDICE \citep{kostrikov2019imitation} for the adversarial IL problem, and OPOLO \citep{zhu2020off} and MobILE \citep{kidambi2021mobile} for the adversarial IfO problem. Another line of research has tackled IfO by directly estimating expert actions and subsequently deploying IL techniques on the estimated state-action pairs \citep{torabi2018behavioral, liu2018imitation, behbahani2019learning, zhang2021learning, zhang2022selfd, shaw2023videodex, yang2023foundation}. Finally, recent studies have investigated offline alternatives to the adversarial IL framework \citep{dadashi2020primal}. 

All of the aforementioned works consider fully observable environments modeled as \emph{Markov Decision Processes (MDPs)}. However, when dealing with pixels, individual observations alone are insufficient for determining optimal actions. As a result, recent works \citep{rafailov2021visual, hu2022model} have treated the V-IL problem as a \emph{Partially Observable Markov Decision Process (POMDP)} \citep{astrom1965optimal}. In particular, \citet{rafailov2021visual} addressed the V-IL problem by proposing a model-based extension \citep{hafner2019dream, hafner2019learning} of generative adversarial IL called VMAIL. The work in \citet{gangwani2020learning} also considered IL in a POMDP in order to handle missing information in the agent state, but did not directly focus on learning from pixels. The more difficult V-IfO problem, on the other hand, has received less attention in the literature. To the best of our knowledge, this problem has only been considered by the recent algorithm PatchAIL \citep{liu2023visual}, where off-policy adversarial IL is performed directly on the pixel space. Different from \citet{liu2023visual}, we first study V-IfO from a theoretical perspective, which motivates an algorithm that performs imitation on a \emph{latent representation of the agent state} rather than directly on the pixel space as in PatchAIL. This difference is crucial to ensure improved computational efficiency. 

Our work is also related to the RL from pixels literature which tackles the challenge of maximizing an agent's expected return end-to-end, from pixels to actions. This approach has proven successful in playing Atari games \citep{mnih2013playing, mnih2015human}. Recently, RL from pixels has also been extended to tackle continuous action space tasks, such as robot locomotion, by leveraging either data augmentation techniques \citep{laskin2020curl, laskin2020reinforcement, kostrikov2020image, lee2020stochastic, raileanu2021automatic, yarats2021mastering} or variational inference \citep{hafner2019dream, hafner2019learning, lee2020stochastic, hafner2020mastering}.

Finally, another line of research has focused on the visual imitation problem in the presence of domain mismatch, also known as third-person imitation learning \citep{stadie2017third, okumura2020domain, cetin2021domain, giammarino2022opportunities}. This paradigm relaxes the assumption that the agent and the expert are defined on the same decision process and represents a generalization of the imitation from experts frameworks introduced in Table~\ref{table_1}.

\section{Preliminaries}
\label{sec:preliminaries}
Unless indicated otherwise, we use uppercase letters (e.g., $S_t$) for random variables, lowercase letters (e.g., $s_t$) for values of random variables, script letters (e.g., $\mathcal{S}$) for sets, and bold lowercase letters (e.g., $\bm{\theta}$) for vectors. Let $[t_1 : t_2]$ be the set of integers $t$ such that $t_1 \leq t \leq t_2$; we write $S_t$ such that $t_1 \leq t \leq t_2$ as $S_{t_1 : t_2}$. We denote with $\mathbb{E}[\cdot]$ expectation, with $\mathbb{P}(\cdot)$ probability, and with $\mathbb{D}_f(\cdot, \cdot)$ an $f$-divergence between two distributions of which the total variation (TV) distance, $\mathbb{D}_{\text{TV}}(\cdot,\cdot)$, and the Jensen-Shannon divergence, $\mathbb{D}_{\text{JS}}(\cdot||\cdot)$, are special cases.

We model the decision process as an infinite-horizon discounted POMDP described by the tuple $(\mathcal{S}, \mathcal{A}, \mathcal{X}, \mathcal{T}, \mathcal{U}, \mathcal{R}, \rho_0, \gamma)$, where $\mathcal{S}$ is the set of states, $\mathcal{A}$ is the set of actions, and $\mathcal{X}$ is the set of observations. $\mathcal{T}:\mathcal{S}\times \mathcal{A} \rightarrow P(\mathcal{S})$ is the transition probability function where $P(\mathcal{S})$ denotes the space of probability distributions over $\mathcal{S}$, $\mathcal{U}:\mathcal{S} \rightarrow P(\mathcal{X})$ is the observation probability function, and $\mathcal{R}:\mathcal{S}\times \mathcal{A} \rightarrow \mathbb{R}$ is the reward function which maps state-action pairs to scalar rewards. Alternatively, the reward function can also be expressed as $\mathcal{R}:\mathcal{S}\times \mathcal{S} \rightarrow \mathbb{R}$ mapping state-transition pairs to scalar rewards rather than state-action pairs. Finally, $\rho_0 \in P(\mathcal{S})$ is the initial state distribution and $\gamma \in [0,1)$ the discount factor. The true environment state $s \in \mathcal{S}$ is unobserved by the agent. Given an action $a\in\mathcal{A}$, the next state is sampled such that $s'\sim\mathcal{T}(\cdot|s,a)$, an observation is generated as $x'\sim\mathcal{U}(\cdot|s')$, and a reward $\mathcal{R}(s,a)$ or $\mathcal{R}(s,s')$ is computed. Note that an MDP is a special case of a POMDP where the underlying state $s$ is directly observed.

\paragraph{Reinforcement learning} Given an MDP and a stationary policy $\pi:\mathcal{S} \to P(\mathcal{A})$, the RL objective is to maximize the expected total discounted return $J(\pi)=\mathbb{E}_{\tau_{\pi}}[\sum_{t=0}^{\infty}\gamma^t \mathcal{R}(s_t,a_t)]$ where $\tau_{\pi} = (s_0,a_0,s_1,a_1,\dots)$. A stationary policy $\pi$ induces a normalized discounted state visitation distribution defined as $d_{\pi}(s) = (1-\gamma)\sum_{t=0}^{\infty}\gamma^t\mathbb{P}(s_t=s | \rho_0, \pi, \mathcal{T})$, and we define the corresponding normalized discounted state-action visitation distribution as $\rho_{\pi}(s,a)=d_{\pi}(s)\pi(a|s)$. Finally, we denote the state value function of $\pi$ as $V^{\pi}(s) = \mathbb{E}_{\tau_{\pi}}[\sum_{t=0}^{\infty}\gamma^t \mathcal{R}(s_t,a_t)|S_0=s]$ and the state-action value function as $Q^{\pi}(s,a) = \mathbb{E}_{\tau_{\pi}}[\sum_{t=0}^{\infty}\gamma^t \mathcal{R}(s_t,a_t)|S_0=s, A_0=a]$. When a function is parameterized with parameters $\bm{\theta} \in \varTheta \subset \mathbb{R}^k$ we write $\pi_{\bm{\theta}}$.

\paragraph{Generative adversarial imitation learning} Assume we have a set of expert demonstrations $\tau_E = (s_{0:T}, a_{0:T})$ generated by the expert policy $\pi_E$, a set of trajectories $\tau_{\bm{\theta}}$ generated by the policy $\pi_{\bm{\theta}}$, and a discriminator network $D_{\bm{\chi}}: \mathcal{S}\times\mathcal{A} \to [0,1]$ parameterized by $\bm{\chi}$. Generative adversarial IL \citep{ho2016generative} optimizes the min-max objective 
\begin{align}
    \min_{\bm{\theta}} \max_{\bm{\chi}} \ \ &\mathbb{E}_{\tau_E}[\log(D_{\bm{\chi}}(s,a))] + \mathbb{E}_{\tau_{\bm{\theta}}}[\log(1 - D_{\bm{\chi}}(s,a))]. \label{eq:IRL_disc}
\end{align}
Maximizing \eqref{eq:IRL_disc} with respect to $\bm{\chi}$ is effectively an inverse RL step where a reward function, in the form of the discriminator $D_{\bm{\chi}}$, is inferred by leveraging $\tau_E$ and $\tau_{\bm{\theta}}$. On the other hand, minimizing \eqref{eq:IRL_disc} with respect to $\bm{\theta}$ can be interpreted as an RL step, where the agent aims to minimize its expected cost. 
It has been demonstrated that optimizing the min-max objective in \eqref{eq:IRL_disc} is equivalent to minimizing $\mathbb{D}_{\text{JS}}(\rho_{\pi_{\bm{\theta}}}(s,a)||\rho_{\pi_E}(s,a))$, so we are recovering the expert state-action visitation distribution \citep{ghasemipour2020divergence}. 


\paragraph{Latent representation in POMDP} When dealing with a POMDP, a policy $\pi_{\bm{\theta}}(x_t)$ that selects an action $a_t$ based on a single observation $x_t \in \mathcal{X}$ is likely to perform poorly since $x_t$ lacks enough information about the unobservable true state $s_t$. It is therefore beneficial to estimate a distribution of the true state from the full history of prior experiences. To that end, we introduce a latent variable $z_t \in \mathcal{Z}$ such that $z_t = \phi(x_{\leq t}, a_{<t})$, where $\phi$ maps the history of observations and actions to $\mathcal{Z}$. Alternatively, when actions are not observable, we have $z_t = \phi(x_{\leq t})$. The latent variable $z_t$ should be estimated such that $\mathbb{P}(s_t|x_{\leq t}, a_{<t}) \approx \mathbb{P}(s_t|z_t)$, meaning that $z_t$ represents a sufficient statistic of the history for estimating a distribution of the unobservable true state $s_t$. It is important to clarify that this does not imply $\mathcal{Z} \equiv \mathcal{S}$.

\section{Theoretical analysis}
\label{sec:theoretical_analysis}
Recall that we consider the V-IfO problem where expert actions are not available and the ground-truth states $s \in \mathcal{S}$ are not observable (see Table~\ref{table_1}). As a result, a latent representation $z \in \mathcal{Z}$ is inferred from the history of observations and used by the learning agent to make decisions. 

Throughout the paper we make the following assumptions: $(i)$ the expert and the agent act on the same POMDP and $(ii)$ the latent variable $z$ can be estimated from the history of observations as $z_t = \phi(x_{\leq t})$ such that $\mathbb{P}(s_t|z_t, a_t) = \mathbb{P}(s_t|z_t) = \mathbb{P}(s_t|x_{\leq t}, a_{<t})$. Assumption $(i)$ is instrumental for both our derivations and experiments. Relaxing this assumption would lead to dynamics mismatch \citep{gangwani2022imitation} and visual domain adaptation problems \citep{giammarino2022opportunities}, which represent interesting extensions for future work. On the other hand, assumption $(ii)$ explicitly states the characteristics required by the latent variable $z$; i.e., $z_t$ can be successfully estimated from the history of observations $x_{\leq t}$ in order to approximate a sufficient statistic of the history. Note that this is a common assumption in the IL literature for POMDPs \citep{gangwani2020learning, rafailov2021visual}, and estimating such a variable is a non-trivial problem that we address in the next section. We further discuss the importance of this assumption from a theoretical perspective in Appendix~\ref{app:auxiliary_results} (Remark~\ref{app:remark_lemma_5}).

On the latent space $\mathcal{Z}$, we can define the normalized discounted latent state visitation distribution as $d_{\pi_{\bm{\theta}}}(z) = (1-\gamma)\sum_{t=0}^{\infty}\gamma^t\mathbb{P}(z_t=z | \rho_0,\pi_{\bm{\theta}},\mathcal{T},\mathcal{U})$ and the normalized discounted latent state-action visitation distribution as $\rho_{\pi_{\bm{\theta}}}(z,a)= d_{\pi_{\bm{\theta}}}(z)\pi_{\bm{\theta}}(a|z)$. Further, we define the latent state-transition visitation distribution as $\rho_{\pi_{\bm{\theta}}}(z,z')=d_{\pi_{\bm{\theta}}}(z)\int_{\mathcal{A}} \mathbb{P}(z'|z,\Bar{a})\pi_{\bm{\theta}}(\Bar{a}|z) d\Bar{a}$ and the normalized discounted joint distribution as $\rho_{\pi_{\bm{\theta}}}(z, a, z') = \rho_{\pi_{\bm{\theta}}}(z,a)\mathbb{P}(z'|z,a)$, where  
\begin{align}
    \begin{split}
        \mathbb{P}(z'|z,a)=& \int_{\mathcal{S}}\int_{\mathcal{S}}\int_{\mathcal{X}}\mathbb{P}(z'|x',a,z)\mathcal{U}(x'|s')\mathcal{T}(s'|s,a)\mathbb{P}(s|z) dx' ds' ds.
    \end{split}
    \label{eq:P_z}
\end{align}
Finally, we obtain $\mathbb{P}_{\pi_{\bm{\theta}}}(a|z, z')$ as
\begin{equation*}
    \mathbb{P}_{\pi_{\bm{\theta}}}(a|z, z') = \frac{\mathbb{P}(z'|z, a)\pi_{\bm{\theta}}(a|z)}{\int_{\mathcal{A}} \mathbb{P}(z'|z,\Bar{a})\pi_{\bm{\theta}}(\Bar{a}|z)d\Bar{a}}.
\end{equation*} 
Note that we write $\mathbb{P}_{\pi_{\bm{\theta}}}$, with $\pi_{\bm{\theta}}$ as subscript, in order to explicitly denote the dependency on the policy and omit the subscript, as in \eqref{eq:P_z}, when such probability depends only on the environment.  

We start by considering the case in which $\mathcal{R}: \mathcal{S} \times \mathcal{A} \to \mathbb{R}$ and $J(\pi)=\mathbb{E}_{\tau_{\pi}}[\sum_{t=0}^{\infty}\gamma^t \mathcal{R}(s_t,a_t)]$. The following Theorem shows how the suboptimality of $\pi_{\bm{\theta}}$ can be upper bounded by the TV distance between latent state-transition visitation distributions, reducing the V-IfO problem to a divergence minimization problem in the latent space $\mathcal{Z}$.
\begin{theorem}
\label{theorem_1}
Consider a POMDP, and let $\mathcal{R}: \mathcal{S} \times \mathcal{A} \to \mathbb{R}$ and $z_t = \phi(x_{\leq t})$ such that $\mathbb{P}(s_t|z_t, a_t) = \mathbb{P}(s_t|z_t) = \mathbb{P}(s_t|x_{\leq t}, a_{<t})$. Then, the following inequality holds: 
\begin{align*}
    \begin{split}
        \big|J(\pi_{E}) - J(\pi_{\bm{\theta}})\big| \leq& \frac{2 R_{\max}}{1-\gamma}\mathbb{D}_{\text{\normalfont{TV}}}\big(\rho_{\pi_{\bm{\theta}}}(z,z'),\rho_{\pi_{E}}(z, z')\big) + C, 
    \end{split}
\end{align*}
where $R_{\max} = \max_{(s,a) \in \mathcal{S} \times \mathcal{A}}|\mathcal{R}(s,a)|$ and 
\begin{equation}
    C = \frac{2 R_{\max}}{1-\gamma}\mathbb{E}_{\rho_{\pi_{\bm{\theta}}}(z, z')}\big[\mathbb{D}_{\text{\normalfont{TV}}}\big(\mathbb{P}_{\pi_{\bm{\theta}}}(a|z, z'),\mathbb{P}_{\pi_{E}}(a|z, z')\big)\big].
    \label{theo_1:C}
\end{equation}
\begin{proof}
Using the definition of $J(\pi_{\bm{\theta}})$, we first upper bound the performance difference between expert and agent by $\mathbb{D}_{\text{\normalfont{TV}}}\big(\rho_{\pi_{\bm{\theta}}}(s, a),\rho_{\pi_{E}}(s, a)\big)$. Next, we bound the latter divergence by $\mathbb{D}_{\text{\normalfont{TV}}}\big(\rho_{\pi_{\bm{\theta}}}(z,a),\rho_{\pi_{E}}(z, a)\big)$ using the assumption $\mathbb{P}(s_t|z_t, a_t) = \mathbb{P}(s_t|z_t)$ and noticing that $ \mathbb{P}(s_t|z_t)$ is policy independent. Finally, we bound this last divergence in terms of $\mathbb{D}_{\text{\normalfont{TV}}}\big(\rho_{\pi_{\bm{\theta}}}(z,z'),\rho_{\pi_{E}}(z, z')\big)$ (Lemma~\ref{lemma_2} in Appendix~\ref{app:auxiliary_results}). We provide the full derivations in Appendix~\ref{app:theo_proofs}.
\end{proof}
\end{theorem}

Theorem~\ref{theorem_1} addresses the challenge of considering rewards that depend on actions without the ability to observe expert actions. Consequently, in our setting, we cannot compute $C$ in \eqref{theo_1:C}. Similar to the MDP case \citep{yang2019imitation}, a sufficient condition for $C=0$ is the injectivity of $\mathbb{P}(z'|z, a)$ in \eqref{eq:P_z} with respect to $a$, indicating that there is only one action corresponding to a given latent state transition. This property ensures that $\mathbb{P}(a|z, z')$ remains unaffected by different executed policies, ultimately reducing $C$ to zero. For the sake of completeness, we formally state this result in Appendix~\ref{app:theo_proofs}. However, in our setting, it is difficult to guarantee the injectivity of $\mathbb{P}(z'|z, a)$ due to its dependence on both the environment through $\mathcal{U}(x'|s')$ and $\mathcal{T}(s'|s,a)$, and the latent variable estimation method through $\mathbb{P}(z'|x',a,z)$ and $\mathbb{P}(s|z)$. Instead, we demonstrate in Theorem~\ref{theorem_2} how redefining the reward function as $\mathcal{R}: \mathcal{S} \times \mathcal{S} \to \mathbb{R}$, which is commonly observed in robotics learning, allows us to reformulate the result in Theorem~\ref{theorem_1} without the additive term $C$ in \eqref{theo_1:C}.

\begin{theorem}
\label{theorem_2}
Consider a POMDP, and let $\mathcal{R}: \mathcal{S} \times \mathcal{S} \to \mathbb{R}$ and $z_t = \phi(x_{\leq t})$ such that $\mathbb{P}(s_t|z_t, a_t) = \mathbb{P}(s_t|z_t) = \mathbb{P}(s_t|x_{\leq t}, a_{<t})$. Then, the following inequality holds:
\begin{align*}
    \begin{split}
        \big|J(\pi_{E}) - J(\pi_{\bm{\theta}})\big| \leq& \frac{2 R_{\max}}{1-\gamma}\mathbb{D}_{\text{\normalfont{TV}}}\big(\rho_{\pi_{\bm{\theta}}}(z,z'),\rho_{\pi_{E}}(z, z')\big),
    \end{split}
\end{align*}
where $R_{\max} = \max_{(s,s') \in \mathcal{S} \times \mathcal{S}}|\mathcal{R}(s,s')|$.
\begin{proof}
    The proof proceeds similarly to the one for Theorem~\ref{theorem_1}, by using that $\mathbb{P}(s,s'|z,z')$ is not characterized by the policy but only by the environment. We show the full proof in Appendix~\ref{app:theo_proofs}.
\end{proof}
\end{theorem}

In summary, Theorems~\ref{theorem_1} and \ref{theorem_2} show that, assuming we have a latent space $\mathcal{Z}$ that can effectively approximate a sufficient statistic of the history, the imitation problem can be performed entirely on this latent space. Note that this is in contrast with the existing literature \citep{liu2023visual}, where imitation is performed on the observation space $\mathcal{X}$. As a result of this analysis, our algorithm is characterized by two main ingredients: a practical method to estimate $z \in \mathcal{Z}$ from sequences of observations, and an efficient optimization pipeline to minimize $\mathbb{D}_{\text{\normalfont{TV}}}\big(\rho_{\pi_{\bm{\theta}}}(z, z'),\rho_{\pi_{E}}(z, z')\big)$.

\section{Latent Adversarial Imitation from Observations}
\label{sec:LAIfO}

In the following, we introduce the main components of our algorithm LAIfO. Motivated by our theoretical analysis in the previous section, our algorithm combines techniques for adversarial imitation from observations and latent variable estimation. First, we outline our adversarial imitation pipeline in the latent space $\mathcal{Z}$, which leverages off-policy adversarial imitation from observations \citep{kostrikov2018discriminator, blonde2019sample, zhu2020off} in order to minimize the divergence between the latent state-transition visitation distributions of the agent and expert. Then, we describe a simple and effective approach for estimating the latent state $z$ that makes use of observations stacking \citep{mnih2013playing, mnih2015human} and data augmentation \citep{laskin2020reinforcement, kostrikov2020image, yarats2021mastering}. Finally, we show how LAIfO can leverage expert videos to enhance the efficiency of RL from pixels in a number of highly challenging tasks. 

\vspace{-0.3cm}
\paragraph{Off-policy adversarial imitation from observations}
Based on the results in Section~\ref{sec:theoretical_analysis}, given a latent variable $z$ that captures a sufficient statistic of the history, we can minimize the suboptimality of the policy $\pi_{\bm{\theta}}$ by solving the minimization problem
\begin{equation}
    \min_{\bm{\theta}} \ \ \ \mathbb{D}_{\text{\normalfont{TV}}}\big(\rho_{\pi_{\bm{\theta}}}(z, z'),\rho_{\pi_{E}}(z, z')\big).
    \label{eq:AIL_min_div}
\end{equation}
We propose to optimize the objective in \eqref{eq:AIL_min_div} using off-policy adversarial IfO. We initialize two replay buffers $\mathcal{B}_E$ and $\mathcal{B}$ to respectively store the sequences of observations generated by the expert and the agent policies, from which we infer the latent state-transitions $(z,z')$. Note that we write $(z,z') \sim \mathcal{B}$ to streamline the notation. Then, given a discriminator $D_{\bm{\chi}}:\mathcal{Z}\times\mathcal{Z} \to [0,1]$, we write
\begin{align}
    \max_{\bm{\chi}} \ \  \mathbb{E}_{(z,z') \sim \mathcal{B}_E}[\log(D_{\bm{\chi}}(z,z'))] + \mathbb{E}_{(z,z') \sim \mathcal{B}}[\log(1 - D_{\bm{\chi}}(z,z'))] + g\big(\nabla_{\bm{\chi}} D_{\bm{\chi}}\big), \label{eq:AIL_BCE} 
\end{align} 

where $g(\cdot)$ is defined in \eqref{eq:penalty_grad_reg}.
As mentioned, alternating the maximization of the loss in \eqref{eq:AIL_BCE} with an RL step leads to the minimization of $\mathbb{D}_{\text{JS}}\big(\rho_{\pi_{\bm{\theta}}}(z, z')||\rho_{\pi_{E}}(z, z')\big)$ \citep{goodfellow2020generative}. Since $\mathbb{D}_{\text{JS}}(\cdot||\cdot)$ can be used to upper bound $\mathbb{D}_{\text{TV}}(\cdot, \cdot)$ (cf. Lemma~\ref{lemma_4} in Appendix~\ref{app:auxiliary_results}), this approach effectively minimizes the loss in \eqref{eq:AIL_min_div}. In order to stabilize the adversarial training process, it is important to ensure local Lipschitz-continuity of the learned reward function \citep{blonde2022lipschitzness}. Therefore, as proposed in \citet{gulrajani2017improved}, we include in \eqref{eq:AIL_BCE} the gradient penalty term
\begin{equation}
    g\big(\nabla_{\bm{\chi}} D_{\bm{\chi}}\big)=\lambda \mathbb{E}_{(\hat{z},\hat{z}') \sim \mathbb{P}_{(\hat{z},\hat{z}')}}[(||\nabla_{\bm{\chi}} D_{\bm{\chi}}(\hat{z},\hat{z}')||_2 - 1)^2],
    \label{eq:penalty_grad_reg}
\end{equation}
where $\lambda$ is a hyperparameter, and $\mathbb{P}_{(\hat{z},\hat{z}')}$ is defined such that $(\hat{z},\hat{z}')$ are sampled uniformly along straight lines between pairs of transitions respectively sampled from $\mathcal{B}_E$ and $\mathcal{B}$. For additional details on the importance of the term in \eqref{eq:penalty_grad_reg} for improved stability, refer to our ablation experiments in Appendix~\ref{app:learning_curves} and to \citet{gulrajani2017improved}. Finally, from a theoretical standpoint, note that we should perform importance sampling correction in order to account for the effect of off-policy data when sampling from $\mathcal{B}$ \citep{queeney2021generalized, queeney2022generalized}. However, neglecting off-policy correction works well in practice and does not compromise the stability of the algorithm \citep{kostrikov2018discriminator}.  

\vspace{-0.3cm}
\paragraph{Latent variable estimation from observations}
Note that the problem in \eqref{eq:AIL_min_div} is defined on the latent space $\mathcal{Z}$. Therefore, we now present a simple and effective method to estimate the latent variable $z$ from sequences of observations. Inspired by the model-free RL from pixels literature, we propose to combine the successful approaches of observations stacking \citep{mnih2013playing, mnih2015human} and data augmentation \citep{laskin2020reinforcement, kostrikov2020image, yarats2021mastering}. We stack together the most recent $d \in \mathbb{N}$ observations, and provide this stack as an input to a feature extractor which is trained during the RL step. More specifically, we define a feature extractor $\phi_{\bm{\delta}}:\mathcal{X}^d\to\mathcal{Z}$ such that $z = \phi_{\bm{\delta}}(x_{t^-:t})$ where $t-t^-+1=d$. When learning from pixels, we also apply data augmentation to the observations stack to improve the quality of the extracted features as in \citet{kostrikov2020image}. We write $\text{aug}(x_{t^-:t})$ to define the augmented stack of observations. The latent representations $z$ and $z'$ are then computed respectively as $z = \phi_{\bm{\delta}}(\text{aug}(x_{t^-:t}))$ and $z' = \phi_{\bm{\delta}}(\text{aug}(x_{t^-+1:t+1}))$. We train the feature extractor $\phi_{\bm{\delta}}$ with the critic networks $Q_{\bm{\psi}_k}$ ($k=1,2$) in order to minimize the loss function
\begin{align}
    \begin{split}
        \mathcal{L}_{\bm{\delta}, \bm{\psi}_k}(\mathcal{B}) &= \mathbb{E}_{(z, a, z')\sim\mathcal{B}}[(Q_{\bm{\psi}_k}(z, a) - y)^2], \\
        y &= r_{\bm{\chi}}(z,z') + \gamma \min_{k={1,2}} Q_{\bar{\bm{\psi}}_k}(z', a'),  
    \end{split} \label{eq:DDPG_critic} \\
    r_{\bm{\chi}}(z, z') &= D_{\bm{\chi}}\big(z, z'\big), \label{eq:AIL_BCE_reward}
\end{align}
where $D_{\bm{\chi}}\big(z, z'\big)$ is the discriminator optimized in \eqref{eq:AIL_BCE}.
In \eqref{eq:DDPG_critic}, $a$ is an action stored in $\mathcal{B}$ used by the agent to interact with the environment, while $a' = \pi_{\bm{\theta}}(z') + \epsilon$ where $\epsilon \sim \text{clip}(\mathcal{N}(0,\sigma^2), -c, c)$ is a clipped exploration noise with $c$ the clipping parameter and $\mathcal{N}(0,\sigma^2)$ a univariate normal distribution with zero mean and $\sigma$ standard deviation. The reward function $r_{\bm{\chi}}(z, z')$ is defined as in \eqref{eq:AIL_BCE_reward}, and $\bar{\bm{\psi}}_1$ and $\bar{\bm{\psi}}_2$ are the slow moving weights for the target Q networks. We provide more implementation details and the complete pseudo-code for our algorithm in Appendix~\ref{app:LAIfO}. 

Note that the feature extractor $\phi_{\bm{\delta}}$ is shared by both the critics $Q_{\bm{\psi}_k}$, the policy $\pi_{\bm{\theta}}$, and the discriminator $D_{\bm{\chi}}$. However, we stop the backpropagation of the gradient from $\pi_{\bm{\theta}}$ and $D_{\bm{\chi}}$ into $\phi_{\bm{\delta}}$. The logic of this choice involves obtaining a latent variable $z$ that is not biased towards any of the players in the adversarial IfO game in \eqref{eq:AIL_BCE}, but only provides the information necessary to determine the expert and agent expected performance. This design is motivated by our theoretical analysis which shows how, provided $\phi_{\bm{\delta}}:\mathcal{X}^d\to\mathcal{Z}$ where $\mathcal{Z}$ approximates a sufficient statistic of the history, all the networks in the adversarial IfO game can be directly defined on $\mathcal{Z}$ as $D_{\bm{\chi}}:\mathcal{Z}\times\mathcal{Z} \to [0,1]$, $Q_{\bm{\psi}_k}:\mathcal{Z}\times\mathcal{A}\to\mathbb{R}$ and $\pi_{\bm{\theta}}:\mathcal{Z} \to P(\mathcal{A})$. This is in contrast with the current literature on V-IfO where all the networks are defined on the observation space $\mathcal{X}$ as $D_{\bm{\chi}}:\mathcal{X}\times\mathcal{X} \to [0,1]$, $Q_{\bm{\psi}_k}:\mathcal{X}\times\mathcal{A}\to\mathbb{R}$ and $\pi_{\bm{\theta}}:\mathcal{X} \to P(\mathcal{A})$.

Finally, note that latent variable estimation is an active research area, and it is possible to apply other techniques such as variational inference~\citep{lee2020stochastic} and contrastive learning~\citep{chen2020simple, grill2020bootstrap}. However, we will show in our experiments that the straightforward approach of observations stacking and data augmentation leads to strong performance in practice, without the need for more complicated estimation procedures. We include an ablation study on the importance of data augmentation in Appendix~\ref{app:learning_curves}.

\vspace{-0.3cm}
\paragraph{Improving RL from pixels using expert videos}
We have so far considered the pure imitation setting where a reward function can only be estimated from expert data. However, for many real-world tasks a simple objective can often be provided to the learning agent. Assuming that videos of experts are also available, we show how we can use LAIfO to accelerate the RL learning process. 

We combine the standard RL objective with our V-IfO objective in \eqref{eq:AIL_min_div}, leading to the combined problem
\begin{align}
\max_{\bm{\theta}} \ \ \ \mathbb{E}_{\tau_{\bm{\theta}}}\left[ \sum_{t=0}^{\infty}\gamma^t \mathcal{R}(s_t,a_t) \right] - \mathbb{D}_{\text{\normalfont{TV}}}\big(\rho_{\pi_{\bm{\theta}}}(z, z'),\rho_{\pi_E}(z, z')\big).
\label{eq:MO-RL}
\end{align}
Using the adversarial IfO pipeline presented in \eqref{eq:AIL_BCE}, we can rewrite \eqref{eq:MO-RL} as 
\begin{align}
    \begin{split}
        \max_{\bm{\theta}} \ \ \ &\mathbb{E}_{\tau_{\bm{\theta}}}\left[\sum_{t=0}^{\infty}\gamma^t \Big( \mathcal{R}(s_t,a_t) + r_{\bm{\chi}}\big(z_t, z_{t+1}\big)\Big)\right],
        \label{eq:MO-RL_discr_IL}
    \end{split}
\end{align}
with $r_{\bm{\chi}}$ in \eqref{eq:AIL_BCE_reward}. By learning $r_{\bm{\chi}}$ with LAIfO and optimizing the problem in \eqref{eq:MO-RL_discr_IL} throughout training, we will show that we are able to significantly improve sample efficiency on challenging humanoid from pixels tasks \citep{tassa2018deepmind} compared to state-of-the-art RL from pixels algorithms \citep{yarats2021mastering}.

\vspace{-0.3cm}
\section{Experiments}
\label{sec:experiments}
In this section, we conduct experiments that aim to answer the following questions: 
\begin{enumerate}
    \item[$(1)$] For the V-IfO problem, how does LAIfO compare to PatchAIL~\citep{liu2023visual}, a state-of-the-art approach for V-IfO, in terms of asymptotic performance and computational efficiency?

    \item[$(2)$] How does the V-IL version of LAIfO with access to expert actions, named \emph{Latent Adversarial Imitation Learning (LAIL)}, compare to VMAIL~\citep{rafailov2021visual}, a state-of-the-art approach for V-IL?

    \item[$(3)$] What is the impact on performance due to partial observability and the absence of expert actions?

    \item[$(4)$] Can LAIfO leverage expert videos to improve the efficiency of RL from pixels in high-dimensional continuous robotic tasks?
\end{enumerate}

For more details about the hardware used to carry out these experiments, all the learning curves, additional ablation studies, and other implementation details, refer to Appendix~\ref{app:learning_curves} and to our \href{https://github.com/VittorioGiammarino/AIL_from_visual_obs/tree/LAIfO}{code}.

\begin{figure}
\centering
    \includegraphics[width=0.9\linewidth]{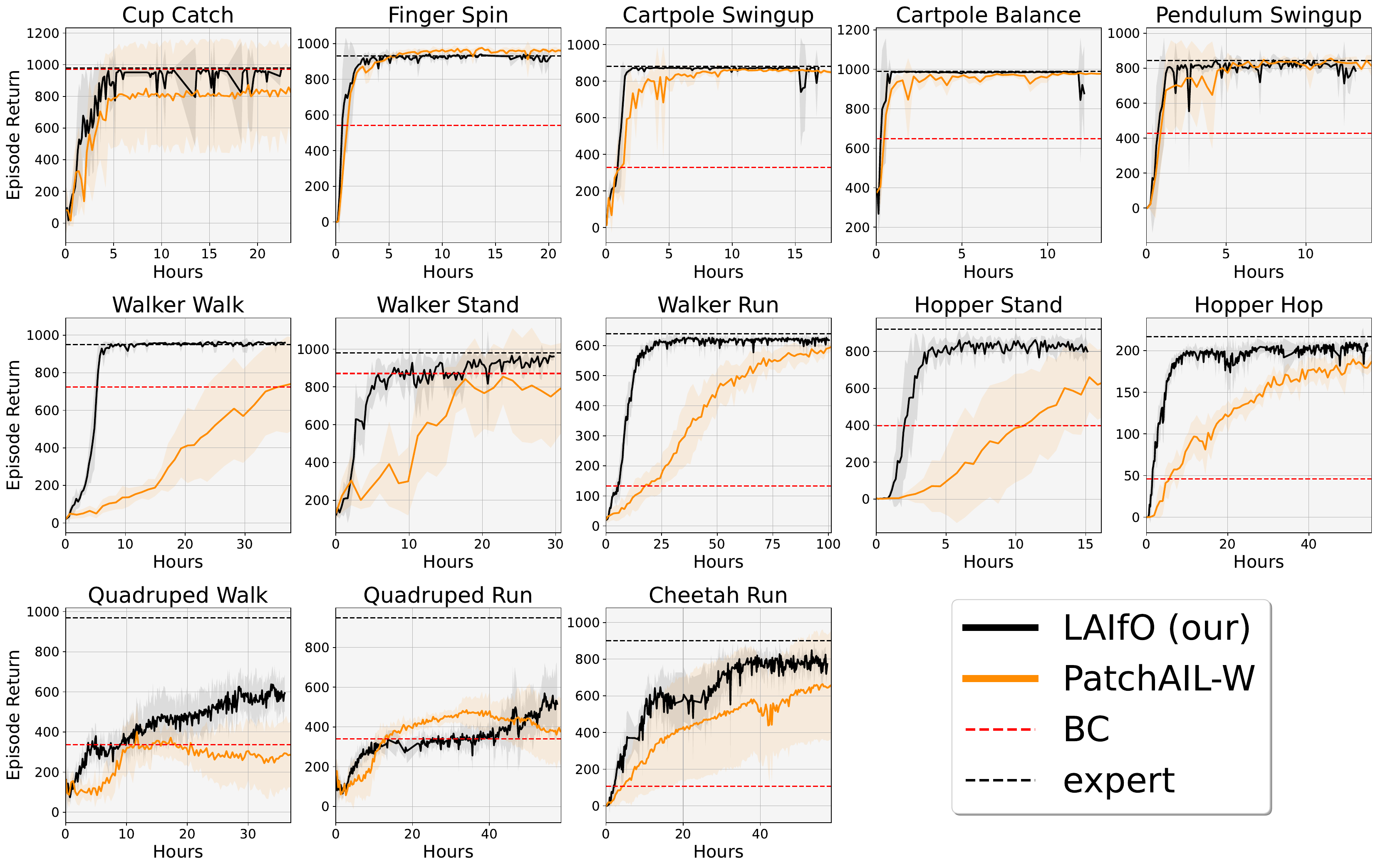}
    \caption{Learning curves for the V-IfO results in Table~\ref{table_V-IfO}. Plots show the average return per episode as a function of wall-clock time. Our algorithm LAIfO achieves state-of-the-art asymptotic performance, and significantly reduces computation time compared to PatchAIL.}
    \label{fig:V-IfO_time}
\end{figure}

\begin{figure}
\centering
    \includegraphics[width=0.9\linewidth]{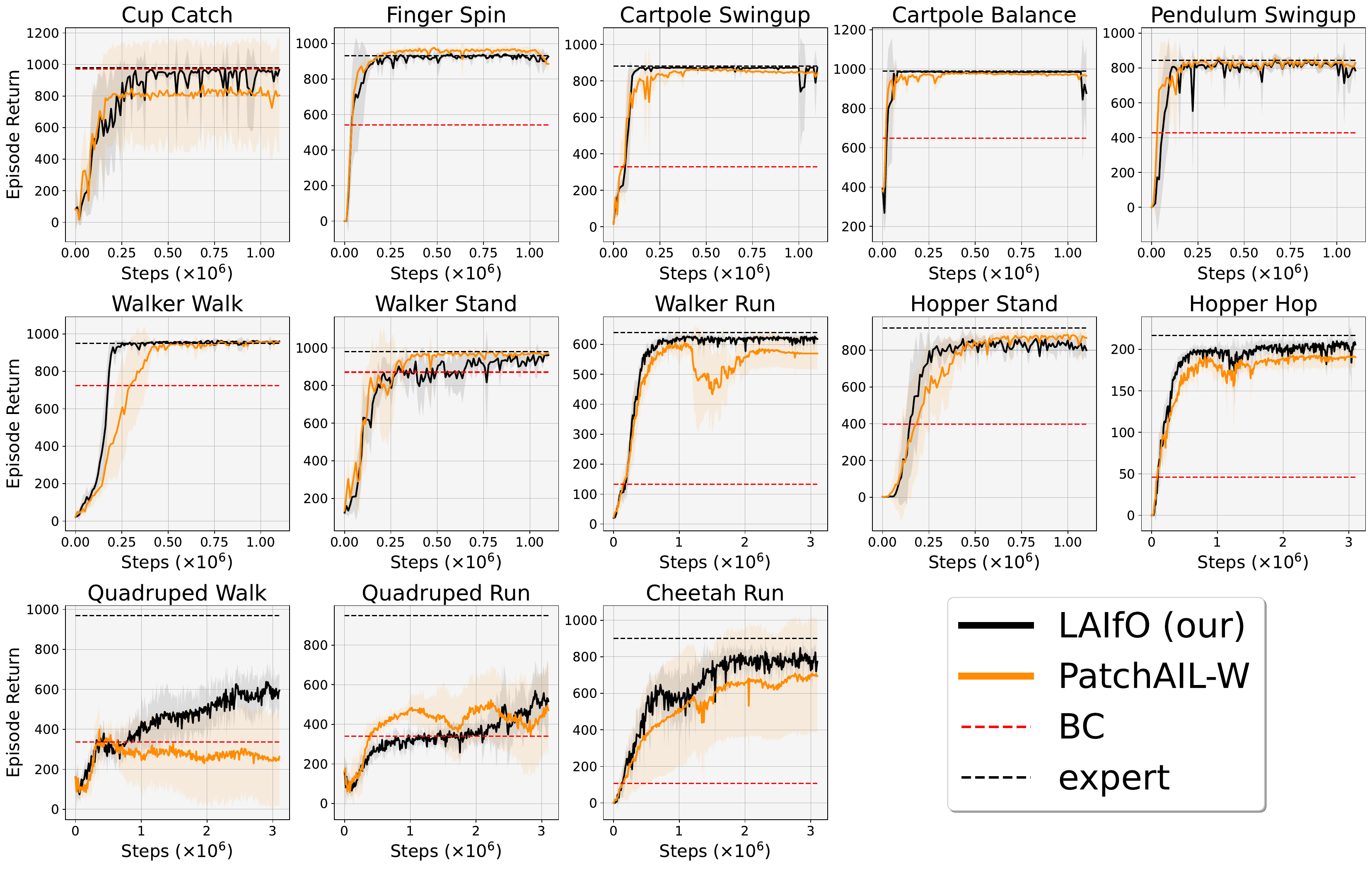}
    \caption{Learning curves for the V-IfO results in Table~\ref{table_V-IfO}. Plots show the average return per episode as a function of training steps.}
    \label{fig:V-IfO_steps}
\end{figure}

\vspace{-0.3cm}
\paragraph{Visual Imitation from Observations} In order to address Question $(1)$, we evaluate LAIfO and PatchAIL~\citep{liu2023visual}, in its weight regularized version denoted by PatchAIL-W, on $13$ different tasks from the DeepMind Control Suite~\citep{tassa2018deepmind}. We also compare these algorithms to Behavioral Cloning (BC)~\citep{pomerleau1988alvinn} for reference. Note that BC uses observation-action pairs. The results are summarized in Table~\ref{table_V-IfO}, Figure~\ref{fig:V-IfO_time}, and Figure~\ref{fig:V-IfO_steps}. Table~\ref{table_V-IfO} includes the asymptotic performance of each algorithm, as well as the ratio of wall-clock times between LAIfO and PatchAIL to achieve 75\% of expert performance. Figure~\ref{fig:V-IfO_time} depicts the average return per episode throughout training as a function of wall-clock time. Moreover, we include in Figure~\ref{fig:V-IfO_steps} plots showing the average return per episode as a function of training steps. These results demonstrate that LAIfO can successfully solve the V-IfO problem, achieving asymptotic performance comparable to the state-of-the-art baseline PatchAIL. Importantly, \emph{LAIfO is significantly more computationally efficient than PatchAIL}. This is well highlighted both in Table~\ref{table_V-IfO} and in Figure~\ref{fig:V-IfO_time}, where we show that \emph{LAIfO always converges faster than PatchAIL in terms of wall-clock time}. This improved computational efficiency is the result of performing imitation on the latent space $\mathcal{Z}$, instead of directly on the high-dimensional observation space $\mathcal{X}$ (i.e., pixel space) as in PatchAIL. Finally, in Table~\ref{table_n_expert} we examine the impact of the amount of expert data on performance. Throughout these experiments, LAIfO does not exhibit a tangible drop in performance due to the decrease of available expert data, and it consistently outperforms PatchAIL.

\begin{table}
    \centering
    \small
    \caption{Experimental results for V-IfO (i.e., imitation from experts with partial observability and without access to expert actions). We use DDPG to train experts in a fully observable setting and collect $100$ episodes of expert data. All of the expert policies can be downloaded by following the instructions in our \href{https://anonymous.4open.science/r/AIL_from_visual_obs-2C3C/README.md}{code} repository. BC is trained offline using expert observation-action pairs for $10^4$ gradient steps. All the other algorithms are trained for $3 \times 10^6$ frames in walker run, hopper hop, cheetah run, quadruped run, and quadruped walk, and $10^6$ frames for the other tasks. We evaluate the learned policies using average performance over $10$ episodes. We run each experiment for $6$ seeds. In the third, fourth and fifth columns, we report mean and standard deviation of final performance over seeds. In the last column, we report the ratio of wall-clock times between LAIfO and PatchAIL to achieve $75\%$ of expert performance. For each task, we \textbf{highlight} the highest asymptotic performance between LAIfO and PatchAIL.}
    \label{table_V-IfO}
    \begin{tabular}{c | c c | c c | c}\toprule
        & Expert & BC & LAIfO (our) & \makecell{PatchAIL-W \\ \citep{liu2023visual}} & \makecell{Wall-clock time ratio \\ to $75\%$ expert performance \\ (LAIfO (our) / PatchAIL-W)} \\ 
        \cmidrule(lr){1-6}
        Cup Catch & $980$ & $971 \pm 9.7$& \bm{$967 \pm 7.6$} & $804 \pm 357$ & $0.69$ \\
        Finger Spin & $932$ & $542 \pm 219$&  \bm{$926 \pm 10.7$} & $885 \pm 30.8$ & $0.67$ \\
        Cartpole Swingup & $881$ & $329 \pm 25.0$& \bm{$873 \pm 3.6$} & $842 \pm 6.7$ & $0.63$ \\
        Cartpole Balance & $990$ & $648 \pm 36.4$& $878 \pm 239$ & \bm{$966 \pm 5.5$} & $0.61$ \\
        Pendulum Swingup & $845$ & $427 \pm 142$& $786 \pm 70.4$ & \bm{$829 \pm 23.7$} & $0.90$ \\
        Walker Walk & $960$ & $723 \pm 137$& \bm{$960 \pm 2.2$} & $955 \pm 7.0$ & $0.15$ \\
        Walker Stand & $980$ & $871 \pm 77.7$& $961 \pm 20.0$ & \bm{$971 \pm 10.5$} & $0.27$ \\
        Walker Run & $640$ & $133 \pm 27.8$& \bm{$618 \pm 4.6$} & $569 \pm 53.2$ & $0.22$ \\
        Hopper Stand & $920$ & $398 \pm 96.4$& $800 \pm 46.7$ & \bm{$867 \pm 33.9$} & $0.16$ \\
        Hopper Hop & $217$ & $45.9 \pm 22.1$& \bm{$206 \pm 8.5$} & $191 \pm 13.0$ & $0.16$ \\
        Quadruped Walk & $970$ & $337 \pm 50.5$& \bm{$594 \pm 92.9$} & $263 \pm 242$ & $\text{NA}^*$ \\
        Quadruped Run & $950$ & $340 \pm 75.2$& \bm{$516 \pm 132$} & $471 \pm 219$ & $\text{NA}^*$ \\
        Cheetah Run & $900$ & $106 \pm 26.3$& \bm{$773 \pm 41.2$} & $695 \pm 312$ & $0.46$ \\
        \bottomrule 
        \multicolumn{5}{l}{\makecell[l]{\small* $75\%$ of the expert performance was not achieved by any algorithm.}}
    \end{tabular}
\end{table}

\begin{figure}
    \centering
    \includegraphics[width=0.9\linewidth]{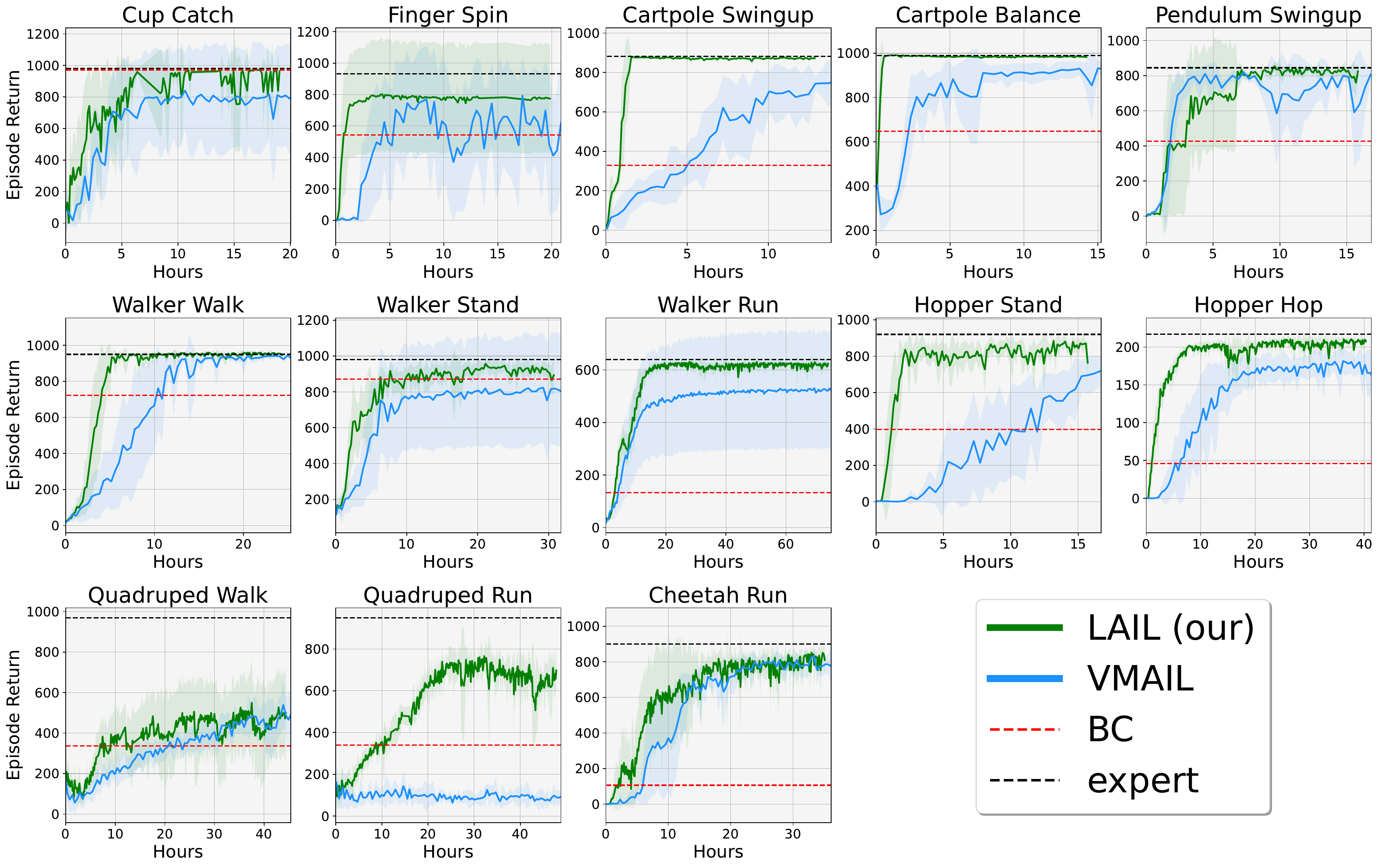}
    \caption{Learning curves for the V-IL results in Table~\ref{table_V-IL}. Plots show the average return per episode as a function of wall-clock time. LAIL outperforms VMAIL in terms of both asymptotic performance and computational efficiency.}
    \label{fig:V-IL_time}
\end{figure}

\begin{figure}
    \centering
    \includegraphics[width=0.9\linewidth]{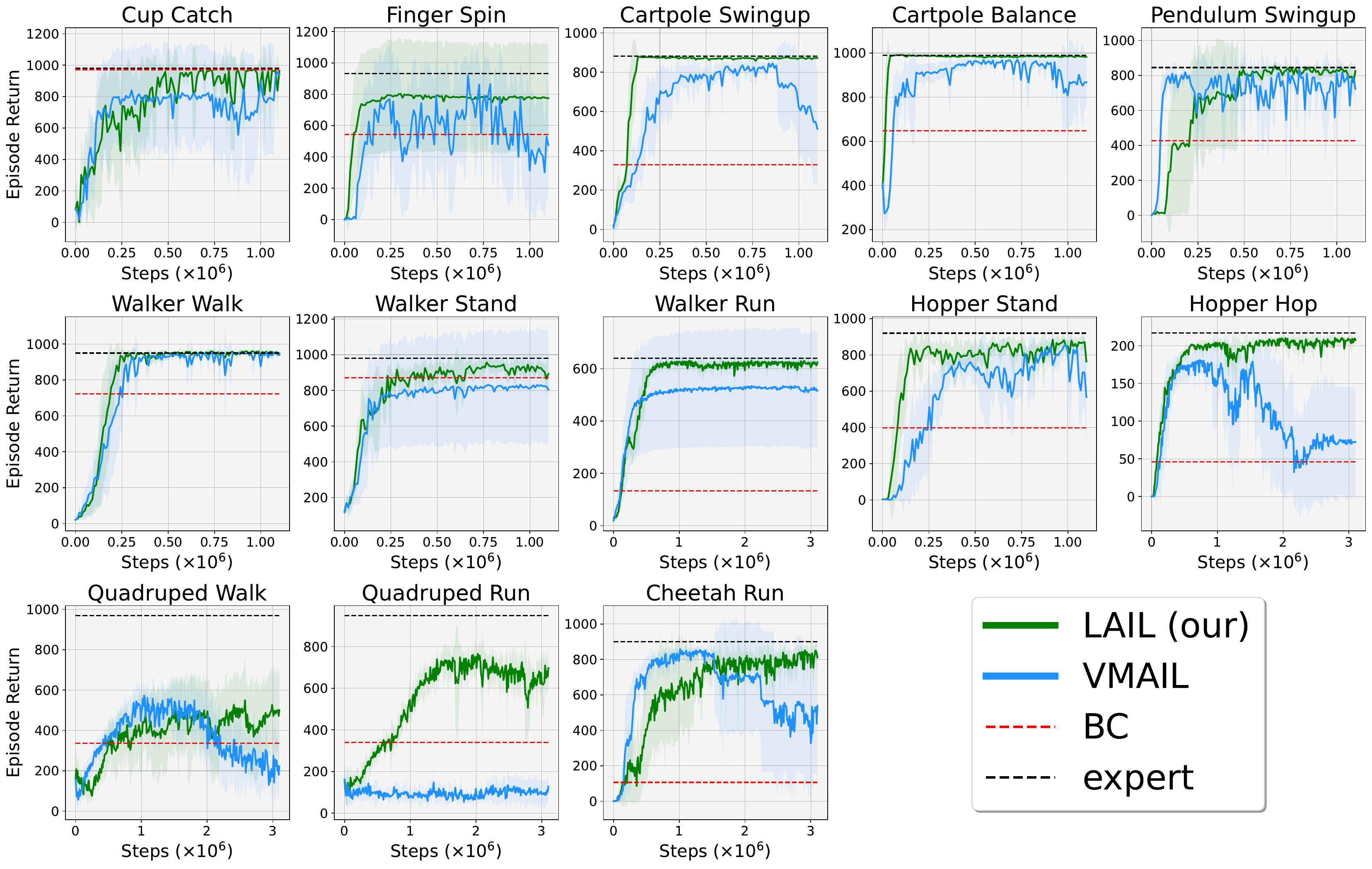}
    \caption{Learning curves for the V-IL results in Table~\ref{table_V-IL}. Plots show the average return per episode as a function of training steps.}
    \label{fig:V-IL_steps}
\end{figure}

\vspace{-0.3cm}
\paragraph{Visual Imitation Learning} To answer Question $(2)$, we test LAIL, the V-IL version of LAIfO, and VMAIL~\citep{rafailov2021visual} using the same experimental setup that we considered in the V-IfO setting. As for V-IfO, we also compare these algorithms to BC for reference. VMAIL stands for \emph{Variational Model Adversarial Imitation Learning}, and represents a model-based version of generative adversarial IL built upon the variational models presented in \citet{hafner2019learning, hafner2019dream, hafner2020mastering}. LAIL is obtained by simply defining the discriminator in \eqref{eq:AIL_BCE} as $D_{\bm{\chi}}:\mathcal{Z}\times\mathcal{A} \to [0,1]$ rather than $D_{\bm{\chi}}:\mathcal{Z}\times\mathcal{Z} \to [0,1]$ as in LAIfO. The results for these experiments are summarized in Table~\ref{table_V-IL}, Figure~\ref{fig:V-IL_time}, and Figure~\ref{fig:V-IL_steps}. Compared to VMAIL, we see that \emph{LAIL achieves better asymptotic performance and better computational efficiency}. While both algorithms perform imitation on a latent space $\mathcal{Z}$, LAIL is a \emph{model-free} algorithm that requires a lower number of learnable parameters compared to the model-based VMAIL. VMAIL must learn an accurate world model during training, which can be a challenging and computationally demanding task. The model learning process contributes to higher wall-clock times, and can also lead to instability in the training process for some environments (cf. Figure~\ref{fig:V-IL_steps}). On the other hand, the model-free approach of LAIL results in stable training that yields faster convergence and better efficiency. Finally, in Table~\ref{table_n_expert} we examine the impact of the amount of expert data on the final results. Throughout these experiments, both LAIL and VMAIL exhibit reliable results and no tangible decrease in performance is observed due to relying on less expert data.

\begin{table}
\centering
\small
\caption{Experimental results for V-IL (i.e., imitation from experts with partial observability and access to expert actions). We use DDPG to train experts in a fully observable setting and collect $100$ episodes of expert data. The experiments are conducted as in Table~\ref{table_V-IfO}. In the third, fourth and fifth columns, we report mean and standard deviation of final performance over seeds. In the last column, we report the ratio of wall-clock times between the two algorithms to achieve $75\%$ of expert performance. For each task, we \textbf{highlight} the highest asymptotic performance between LAIL and VMAIL.}
\label{table_V-IL}
    \begin{tabular}{c | c c | c c | c}\toprule
        & Expert & BC & LAIL (our) & \makecell{VMAIL \\ \citep{rafailov2021visual}} & \makecell{Wall-clock time ratio \\ to $75\%$ expert performance \\ (LAIL (our) / VMAIL)} \\ 
        \cmidrule(lr){1-6}
        Cup Catch & $980$ & $971 \pm 9.7$ & \bm{$962 \pm 18.0$} & $939 \pm 37.4$ & $0.31$ \\
        Finger Spin & $932$ & $542 \pm 219$ & \bm{$775 \pm 345$} & $476 \pm 352$ & $0.21$ \\
        Cartpole Swingup & $881$ & $329 \pm 25.0$ & \bm{$873 \pm 3.0$} & $512 \pm 291$ & $0.13$ \\
        Cartpole Balance & $990$ & $648 \pm 36.3$ & \bm{$982 \pm 1.6$} & $868 \pm 104$ & $0.10$ \\
        Pendulum Swingup & $845$ & $427 \pm 142$ & \bm{$825 \pm 45.1$} & $723 \pm 143$ & $1.43$ \\
        Walker Walk & $960$ & $723 \pm 137$ & \bm{$946 \pm 8.5$} & $939 \pm 9.8$ & $0.40$ \\
        Walker Stand & $980$ & $871 \pm 77.7$ & \bm{$893 \pm 106$} & $805 \pm 309$ & $0.82$ \\
        Walker Run & $640$ & $133 \pm 27.8$ & \bm{$625 \pm 5.1$} & $516 \pm 224$ & $0.58$ \\
        Hopper Stand & $920$ & $398 \pm 96.4$ & \bm{$764 \pm 111$} & $567 \pm 285$ & $0.12$ \\
        Hopper Hop & $217$ & $45.9 \pm 22.1$ &\bm{$208 \pm 3.1$} & $72.3 \pm 73.0$ & $0.23$ \\
        Quadruped Walk & $970$ & $337 \pm 50.5$ & \bm{$500 \pm 182$} & $223 \pm 51.9$ & $\text{NA}^*$ \\
        Quadruped Run & $950$ & $340 \pm 75.2$ & \bm{$697 \pm 102$} & $127 \pm 66.0$ & $\text{NA}^{**}$ \\
        Cheetah Run & $900$ & $106 \pm 26.3$ &\bm{$811 \pm 67.9$} & $539 \pm 367$ & $0.83$ \\
        \bottomrule
        \multicolumn{5}{l}{\makecell[l]{\small* $75\%$ of the expert performance was not achieved by any algorithm.}} \\
        \multicolumn{5}{l}{\makecell[l]{\small** $75\%$ of the expert performance was not achieved by VMAIL.}}
    \end{tabular}
\end{table}

\begin{table}[h!]
\centering
\small
\caption{Performance for different numbers of expert episodes on Walker Run. Except for the number of expert episodes, the experiments are conducted as in Table~\ref{table_V-IfO} and Table~\ref{table_V-IL}. We report mean and standard deviation after training for $10^6$ frames. For both the V-IL and the V-IfO settings, we \textbf{highlight} the highest performance.}
\label{table_n_expert}
\begin{tabular}{c c c c c c}\toprule
\multirow{2}{*}{Walker Run}  & \multicolumn{5}{c}{n Expert episodes}  \\ 
\cmidrule(lr){2-6}
& $1$ & $10$ & $20$ & $50$ & $100$ \\
\cmidrule(lr){1-6}
BC & $28.6 \pm 1.9$ & $70.7 \pm 1.8$ & $87.0 \pm 1.8$ & $113 \pm 1.8$ & $133 \pm 1.8$ \\ 
VMAIL~\citep{rafailov2021visual} & $520 \pm 224$ & \bm{$630 \pm 21.9$} & $517 \pm 221$ & \bm{$619 \pm 11.8$} & $524 \pm 222$ \\
LAIL (our) & \bm{$614 \pm 7.6$} & $626 \pm 7.3$ & \bm{$627 \pm 6.8$} & $611 \pm 22.5$ & \bm{$613 \pm 20.3$} \\
\cmidrule(lr){1-6}
PatchAIL-W~\citep{liu2023visual} & $494 \pm 199$ & $561 \pm 36.0$ & $586 \pm 39.7$ & $604 \pm 22.5$ & $598 \pm 34.7$ \\
LAIfO (our) & \bm{$612 \pm 22.8$} & \bm{$620 \pm 6.4$} & \bm{$623 \pm 7.6$} & \bm{$618 \pm 10.1$} & \bm{$621 \pm 4.8$} \\
\bottomrule
\end{tabular}
\end{table}

\begin{figure}[t]
    \centering
    \includegraphics[width=0.7\linewidth]{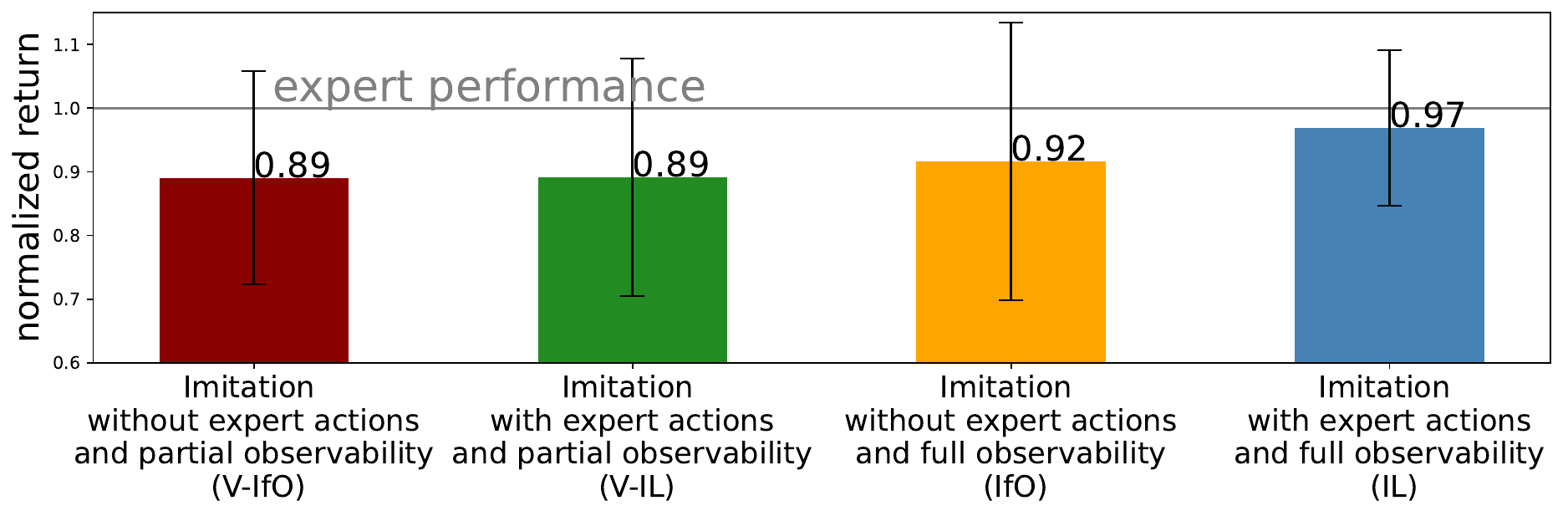}
    \caption{Normalized returns obtained by each type of imitation from experts over the tasks in Table~\ref{table_V-IfO}. For each task and random seed, normalized returns represent performance divided by the expert performance for the considered task (Expert column in Table~\ref{table_V-IfO} and Table~\ref{table_V-IL}). For each type of imitation from experts, we plot mean and standard deviation over the full set of runs. The performance of our algorithms in the partially observable setting are comparable to the performance in the fully observable setting, and the absence of expert actions and partial observability leads only to marginal performance degradation.}
    \label{fig:cum_hist}
\end{figure}

\vspace{-0.3cm}
\paragraph{Ablation study} In order to answer Question $(3)$, we compare performance for each type of imitation from experts in Table~\ref{table_1}. For the partially observable setting, we consider our algorithms LAIL and LAIfO. For the fully observable setting, we consider DAC~\citep{kostrikov2018discriminator} and our implementation of \emph{DAC from Observations (DACfO)}. We provide the full learning curves for DAC and DACfO in Appendix~\ref{app:learning_curves} (cf. Table~\ref{table_IL-IfO} and Figure~\ref{fig:MDP_app}). The results are summarized in Figure~\ref{fig:cum_hist}, which shows the average normalized return obtained by each algorithm throughout the different tasks in Table~\ref{table_V-IfO}. These experiments highlight how our algorithms can successfully address the absence of expert actions and partial observability, suffering only marginal performance degradation due to these additional challenges. As explained in our theoretical analysis in Section~\ref{sec:theoretical_analysis}, partial observability is addressed by estimating a latent state representation that successfully approximates a sufficient statistic of the history. On the other hand, marginal degradation due to the absence of expert actions occurs either because we are in the context described by Theorem~\ref{theorem_2}, where the environment reward function does not depend on actions, or because $C$ in Theorem~\ref{theorem_1} becomes negligible.

\begin{figure}[h!]
    \centering
    \includegraphics[width=0.9\linewidth]{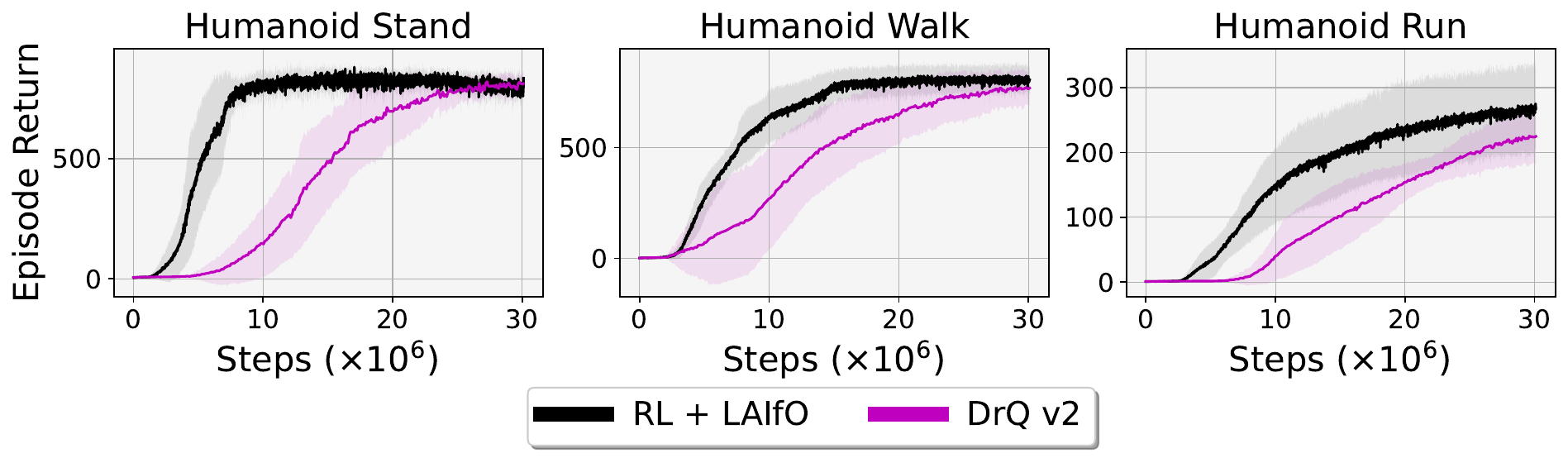}
    \caption{Performance using the multi-objective RL framework in \eqref{eq:MO-RL} on the humanoid environment. The experiments are designed as in Table~\ref{table_V-IfO}. We report mean and standard error over seeds. For DrQv2, note that we use as comparison the curves provided in the official \href{https://github.com/facebookresearch/drqv2}{Github repository} of the paper.}
    \label{fig:RS}
\end{figure}

\paragraph{Improving RL using expert videos} We answer Question $(4)$ by applying LAIfO to the problem in \eqref{eq:MO-RL} for the humanoid from pixels environment. We consider the state-of-the-art model-free RL from pixels algorithms, DrQv2~\citep{yarats2021mastering} as a baseline. The results are illustrated in Figure~\ref{fig:RS}. By leveraging expert videos, we see that our algorithm significantly outperforms the baseline in terms of sample efficiency. This result highlights the value of leveraging expert videos to improve the sample efficiency of RL algorithms, in particular when dealing with challenging tasks.

\section{Conclusion}
\label{sec:conclusion}
In this work, we formally analyzed the V-IfO problem and introduced our algorithm LAIfO as an effective solution. We experimentally showed that our approach matches the performance of state-of-the-art V-IL and V-IfO methods, while requiring significantly less computational effort due to our model-free approach in latent space. Furthermore, we showed how LAIfO can be used to improve the efficiency and asymptotic performance of RL methods by leveraging expert videos.

\paragraph{Limitations and future work} Despite the advancement in addressing the V-IfO problem, it is important to understand the limitations of our approach. The primary limitation arises from the assumption that the expert and the agent act within the same POMDP. In realistic scenarios, such alignment rarely occurs, emphasizing the need for methods that can handle dynamics mismatch and visual domain adaptation. This is a crucial next step towards enabling successful learning from expert videos. Furthermore, throughout this work we have used adversarial learning for divergence minimization between distributions. Adversarial learning can introduce optimization challenges and stability issues. While we propose practical solutions to mitigate these problems, exploring alternatives to this framework offers another interesting avenue for future research. Additionally, from an experimental standpoint, our emphasis has been on robotics control tasks. In the future, we plan to address navigation tasks, considering not only third-view perspectives but also egocentric camera viewpoints. In this context, a challenging and relevant consideration is the correspondence problem, i.e., the problem of enabling egocentric policy learning directly from third-view videos of experts. Finally, we are interested in testing our algorithms on more realistic scenarios that go beyond simulated environments. Our long-term goal is to provide solutions for real-world problems, such as vehicle navigation and robotic manipulation, and to enable imitation directly from videos of biological systems such as humans and animals.

\section{Acknowledgements}

Vittorio Giammarino and Ioannis Ch. Paschalidis were partially supported by the NSF under grants IIS-
1914792, CCF-2200052, and  ECCS-2317079, by the ONR under grant N00014-19-1-2571, by the DOE under 
grant DE-AC02-05CH11231, by the NIH under grant UL54 TR004130, and by Boston University. James Queeney was exclusively supported by Mitsubishi Electric Research Laboratories.

\bibliography{mybib.bib}
\bibliographystyle{tmlr}

\clearpage
\appendix

\section{Broader impacts}

Learning from expert videos will have a transformative impact on robotics, revolutionizing the way robots acquire skills, collaborate with humans, and operate in various domains. This technology holds the potential to enhance efficiency and accessibility across industries, while also fostering human-robot collaboration.

Despite the numerous positive implications, it is important to consider potential negative aspects that may arise from this technology. As in all data-driven methods, biases in the data should always be a cause for concern. Expert observations may inadvertently contain biases, reflecting the preferences or limitations of the demonstrators. If these biases are not properly addressed, robots could potentially perpetuate and amplify those biases in their actions, leading to unsafe or unfair behaviors.

We believe it is crucial to address these potential negative implications through careful design, robust validation, and ongoing research. Responsible development and deployment of machine learning in robotics should consider ethical considerations, promote fairness and transparency, and ensure the technology benefits society as a whole.

\section{Auxiliary results}
\label{app:auxiliary_results}
In the following, we introduce a series of auxiliary definitions and results which are then used to prove Theorems~\ref{theorem_1} and \ref{theorem_2}. 
\begin{definition}[$f$-divergence]
    Let $P$ and $Q$ be two probability distributions over a measurable space $(\Omega, \mathcal{F})$, such that $P$ is absolutely continuous with respect to $Q$. Then, for a convex function $f:[0,\infty)\to (-\infty, \infty]$ such that $f(x)$ is finite for all $x>0$, $f(1)=0$, and $f(0)=\lim_{t\to 0^{+}}f(t)$, the $f$-divergence is defined as  
    \begin{equation}
        \mathbb{D}_f(P||Q) \triangleq \mathbb{E}_Q\bigg[f\bigg(\frac{P}{Q}\bigg)\bigg],
    \end{equation}
    where $P = \frac{dP}{d\mu}$ and $Q=\frac{dQ}{d\mu}$ with $\mu$ any dominating probability measure \citep{polyanskiy2014lecture}. 
    \label{def:f_div}
\end{definition}
From Definition~\ref{def:f_div}, the following $f$-divergences can be derived:

\begin{itemize}
    \item Total variation distance: $f(x)=\frac{1}{2}|x-1|$,
        \begin{equation}
            \mathbb{D}_{\text{TV}}(P,Q) \triangleq \frac{1}{2} \mathbb{E}_Q\bigg[ \bigg| \frac{P}{Q}-1\bigg|\bigg] = \frac{1}{2}\int|P - Q| d\mu.
        \end{equation}

    \item Jensen-Shannon divergence: $f(x) = x\log\frac{2x}{x+1} + \log \frac{2}{x+1}$,
    \begin{equation}
        \mathbb{D}_{\text{JS}}(P||Q) \triangleq  \mathbb{D}_{\text{KL}}\Big(P\Big|\Big|\frac{P+Q}{2}\Big) + \mathbb{D}_{\text{KL}}\Big(Q\Big|\Big|\frac{P+Q}{2}\Big).
        \label{eq:DJS}
    \end{equation}
\end{itemize}

\begin{lemma}[Relation between $f$-divergences]
The following inequality holds:
\label{lemma_4}
\begin{equation}
    \mathbb{D}_{\text{\normalfont{TV}}}(P,Q) \leq \sqrt{\mathbb{D}_{\text{\normalfont{JS}}}(P||Q)}.
    \label{eq:lemma1_part2}
\end{equation}
\begin{proof}

For \eqref{eq:lemma1_part2}, let $M=\frac{P+Q}{2}$. Note that
\begin{align}
    2\mathbb{D}_{\text{\normalfont{TV}}}(P,M) =& \int|P - M| d\mu \nonumber\\
    =& \int\Big|P -\Big(\frac{P+Q}{2}\Big)\Big|d\mu \nonumber\\
    =& \frac{1}{2}\int|P - Q|d\mu \nonumber\\
    =& \mathbb{D}_{\text{\normalfont{TV}}}(P,Q). \nonumber
\end{align}

Similarly, $2\mathbb{D}_{\text{\normalfont{TV}}}(Q,M) = \mathbb{D}_{\text{\normalfont{TV}}}(P,Q)$. Therefore,
\begin{align}
    (\mathbb{D}_{\text{TV}}(P,Q))^2 =&  \frac{1}{2}(\mathbb{D}_{\text{TV}}(P,Q))^2 + \frac{1}{2}(\mathbb{D}_{\text{TV}}(P,Q))^2 \nonumber\\
    =& 2(\mathbb{D}_{\text{TV}}(P,M))^2 + 2(\mathbb{D}_{\text{TV}}(Q,M))^2 \nonumber\\
    \leq& \mathbb{D}_{\text{KL}}(P||M) + \mathbb{D}_{\text{KL}}(Q||M) \label{eq:lemma1_part2_step1} \\
    =& \mathbb{D}_{\text{JS}}(P||Q), \label{eq:lemma1_part2_step2}
\end{align}
where in \eqref{eq:lemma1_part2_step1} we used Pinsker's inequality and in \eqref{eq:lemma1_part2_step2} the definition of Jensen-Shannon divergence from \eqref{eq:DJS}. The result follows by taking the square root of both sides.
\end{proof}
\end{lemma}

\begin{lemma}
\label{lemma_1}
Consider a POMDP, under the assumption that the agent $\pi_{\bm{\theta}}$ and the expert $\pi_E$ act on the same POMDP and with $\mathbb{P}(z'|z,a)$ defined in \eqref{eq:P_z}. Then, the following holds:
\begin{align}
\begin{split}
    &\mathbb{D}_{f}(\rho_{\pi_{\bm{\theta}}}(z, a, z'),\rho_{\pi_E}(z, a, z'))= \mathbb{D}_{f}(\rho_{\pi_{\bm{\theta}}}(z, a),\rho_{\pi_E}(z, a)).
\end{split}
\end{align}
\begin{proof}
We have that
    \begin{align}
        \mathbb{D}_{f}(\rho_{\pi_{\bm{\theta}}}(z, a, z'),\rho_{\pi_E}(z, a, z')) =& \mathbb{E}_{(z,a,z')\sim \rho_{\pi_E}(z,a,z')}\bigg[f\bigg(\frac{\rho_{\pi_{\bm{\theta}}}(z, a, z')}{\rho_{\pi_E}(z, a, z')}\bigg)\bigg] \nonumber \\
        =&\mathbb{E}_{(z,a,z')\sim\rho_{\pi_E}(z,a)\mathbb{P}(z'|z,a)}\bigg[f\bigg( \frac{\rho_{\pi_{\bm{\theta}}}(z,a)\mathbb{P}(z'|z,a)}{\rho_{\pi_E}(z,a)\mathbb{P}(z'|z,a)}\bigg)\bigg] \label{lemma_1:step} \\ 
        =&\mathbb{E}_{(z,a)\sim \rho_{\pi_E}(z,a)}\bigg[f\bigg( \frac{\rho_{\pi_{\bm{\theta}}}(z,a)}{\rho_{\pi_E}(z,a)}\bigg)\bigg]  \nonumber \\ 
        =&\mathbb{D}_{f}(\rho_{\pi_{\bm{\theta}}}(z, a),\rho_{\pi_E}(z, a)). \nonumber
    \end{align}
     This result was similarly proved for fully observable MDPs in \citet{yang2019imitation}. We generalize it for the POMDP case using the definition of $\mathbb{P}(z'|z,a)$ in \eqref{eq:P_z}, which does not depend on any policy but only on the environment. By assumption, expert and agent act on the same POMDP and this yields the step in \eqref{lemma_1:step} from which the proof follows. 
\end{proof}
\end{lemma}

\begin{lemma}
\label{lemma_2}
Consider a POMDP, under the assumption that the agent $\pi_{\bm{\theta}}$ and the expert $\pi_E$ act on the same POMDP. Then, the following holds:
\begin{align}
    \begin{split}
        \mathbb{D}_{\text{\normalfont{TV}}}(\rho_{\pi_{\bm{\theta}}}(z, a),\rho_{\pi_E}(z, a)) \leq& \ \ \mathbb{E}_{(z,z')\sim\rho_{\pi_{\bm{\theta}}}(z,z')}\bigg[\mathbb{D}_{\text{\normalfont{TV}}}\big(\mathbb{P}_{\pi_{\bm{\theta}}}(a| z, z'),\mathbb{P}_{\pi_E}(a| z, z')\big)\bigg] \\ &+ \mathbb{D}_{\text{\normalfont{TV}}}(\rho_{\pi_{\bm{\theta}}}(z, z'),\rho_{\pi_E}(z, z')).
    \end{split}
\end{align}
\begin{proof}
    From Lemma~\ref{lemma_1}, we can write $\mathbb{D}_{\text{TV}}(\rho_{\pi_{\bm{\theta}}}(z, a),\rho_{\pi_E}(z, a)) = \mathbb{D}_{\text{TV}}(\rho_{\pi_{\bm{\theta}}}(z, a, z'),\rho_{\pi_E}(z, a, z'))$. Then,
    \begin{align}
        \mathbb{D}_{\text{TV}}&(\rho_{\pi_{\bm{\theta}}}(z, a, z'),\rho_{\pi_E}(z, a, z')) \nonumber \\
        &= \frac{1}{2}\int_{\mathcal{Z}}\int_{\mathcal{Z}}\int_{\mathcal{A}}\big|\rho_{\pi_{\bm{\theta}}}(z, a, z') - \rho_{\pi_E}(z, a, z')\big| dadz'dz \nonumber \\
        &= \frac{1}{2}\int_{\mathcal{Z}}\int_{\mathcal{Z}}\int_{\mathcal{A}}\big|\rho_{\pi_{\bm{\theta}}}(z, z')\mathbb{P}_{\pi_{\bm{\theta}}}(a|z,z') - \rho_{\pi_{E}}(z, z')\mathbb{P}_{\pi_E}(a|z,z')\big| dadz'dz \nonumber \\
        &= \frac{1}{2}\int_{\mathcal{Z}}\int_{\mathcal{Z}}\int_{\mathcal{A}}\big|\rho_{\pi_{\bm{\theta}}}(z, z')\mathbb{P}_{\pi_{\bm{\theta}}}(a|z,z') - \rho_{\pi_{\bm{\theta}}}(z, z')\mathbb{P}_{\pi_E}(a|z,z') \nonumber \\
        &\ \ \ \qquad \qquad \qquad + \rho_{\pi_{\bm{\theta}}}(z, z')\mathbb{P}_{\pi_E}(a|z,z') - \rho_{\pi_{E}}(z, z')\mathbb{P}_{\pi_E}(a|z,z')\big| dadz'dz  \nonumber \\
        \begin{split}
        &\leq \frac{1}{2}\int_{\mathcal{Z}}\int_{\mathcal{Z}}\int_{\mathcal{A}}\big|\rho_{\pi_{\bm{\theta}}}(z, z')\mathbb{P}_{\pi_{\bm{\theta}}}(a|z,z') - \rho_{\pi_{\bm{\theta}}}(z, z')\mathbb{P}_{\pi_E}(a|z,z')\big| dadz'dz \\ 
        &\ \ \ + \frac{1}{2}\int_{\mathcal{Z}}\int_{\mathcal{Z}}\int_{\mathcal{A}}\big|\rho_{\pi_{\bm{\theta}}}(z, z')\mathbb{P}_{\pi_E}(a|z,z') - \rho_{\pi_{E}}(z, z')\mathbb{P}_{\pi_E}(a|z,z')\big| dadz'dz  \label{app:lemma4_triangle_ineq_step}
        \end{split}\\
        &= \mathbb{E}_{(z,z')\sim\rho_{\pi_{\bm{\theta}}}(z,z')}\bigg[\mathbb{D}_{\text{\normalfont{TV}}}\big(\mathbb{P}_{\pi_{\bm{\theta}}}(a| z, z'),\mathbb{P}_{\pi_E}(a| z, z')\big)\bigg] + \mathbb{D}_{\text{\normalfont{TV}}}(\rho_{\pi_{\bm{\theta}}}(z, z'),\rho_{\pi_E}(z, z')), \nonumber
    \end{align}
    where \eqref{app:lemma4_triangle_ineq_step} follows from the triangle inequality.
\end{proof}
\end{lemma}

\begin{lemma}
    \label{lemma_3}
    Consider a POMDP and let $z_t$ be a latent state representation such that $\mathbb{P}(s_t|z_t, a_t) = \mathbb{P}(s_t|z_t) = \mathbb{P}(s_t|x_{\leq t}, a_{<t})$. Then, the filtering posterior distributions $\mathbb{P}(s_t|z_t)$ and $\mathbb{P}(s_{t+1},s_t|z_{t+1},z_t)$ do not depend on the policy $\pi$ but only on the environment. 
    \begin{proof}
        The proof follows by applying Bayes rule and considering the definition of the latent variable $z_t$. We start from $\mathbb{P}(s_t|z_t)$:
        \begin{align}
            \mathbb{P}(s_t|z_t) =& \mathbb{P}(s_t|x_t, a_{t-1}, z_{t-1}) \nonumber \\ 
            =& \frac{\mathbb{P}(x_t|s_t, a_{t-1}, z_{t-1})\mathbb{P}(s_t| a_{t-1}, z_{t-1})}{\mathbb{P}(x_t|a_{t-1}, z_{t-1})} \nonumber \\
            =& \frac{\mathcal{U}(x_t|s_t)\int_{\mathcal{S}}\mathcal{T}(s_t|s_{t-1},a_{t-1}) \mathbb{P}(s_{t-1}|z_{t-1}) ds_{t-1}}{\int_{\mathcal{S}}\int_{\mathcal{S}}\mathcal{U}(x_t|s_t)\mathcal{T}(s_t|s_{t-1},a_{t-1}) \mathbb{P}(s_{t-1}|z_{t-1}) ds_t ds_{t-1}}, \nonumber
        \end{align}
        where the denominator can be seen as a normalizing factor. Note that there is no dependence on the policy $\pi$. Similarly, for $\mathbb{P}(s_{t+1},s_t|z_t,z_{t+1})$ we have that
        \begin{align}
            \mathbb{P}(s_{t+1},s_t|z_t,z_{t+1}) =& \mathbb{P}(s_t, s_{t+1}|x_{t+1}, a_t, z_t) \nonumber \\
            =& \frac{\mathbb{P}(x_{t+1}|s_t, s_{t+1}, a_t, z_t)\mathbb{P}(s_t, s_{t+1}|a_t, z_t)}{\mathbb{P}(x_{t+1}|a_t, z_t)} \nonumber \\
            =& \frac{\mathcal{U}(x_{t+1}|s_{t+1})\mathcal{T}(s_{t+1}|s_{t},a_{t}) \mathbb{P}(s_{t}|z_{t})}{\int_{\mathcal{S}}\int_{\mathcal{S}}\mathcal{U}(x_{t+1}|s_{t+1})\mathcal{T}(s_{t+1}|s_{t},a_{t}) \mathbb{P}(s_{t}|z_{t}) ds_{t+1} ds_{t}}, \nonumber
        \end{align}
        which does not depend on $\pi$ since $\mathbb{P}(s_t|z_t)$ does not depend on $\pi$.
    \end{proof}
\end{lemma}

\begin{remark}[Remark on the assumptions in Lemma~\ref{lemma_3}]
\label{app:remark_lemma_5}
The assumption $\mathbb{P}(s_t|z_t) = \mathbb{P}(s_t|x_{\leq t}, a_{<t})$ in Lemma~\ref{lemma_3} is crucial to prove the independence of $\mathbb{P}(s_t|z_t)$ on the policy $\pi$. Note that by only considering $\mathbb{P}(s_t|z_t) = \mathbb{P}(s_t|x_{\leq t})$, we have
        \begin{align}
            \mathbb{P}(s_t|z_t) =& \mathbb{P}(s_t|x_t, z_{t-1}) \nonumber \\ 
            =& \frac{\mathbb{P}(x_t|s_t, z_{t-1})\mathbb{P}(s_t| z_{t-1})}{\mathbb{P}(x_t| z_{t-1})} \nonumber \\
            =& \frac{\mathcal{U}(x_t|s_t)\int_{\mathcal{A}}\int_{\mathcal{S}}\mathcal{T}(s_t|s_{t-1},a_{t-1})\pi(a_{t-1}|z_{t-1})\mathbb{P}(s_{t-1}|z_{t-1}) ds_{t-1} da_{t-1}}{\int_{\mathcal{S}}\int_{\mathcal{A}}\int_{\mathcal{S}}\mathcal{U}(x_t|s_t)\mathcal{T}(s_t|s_{t-1},a_{t-1})\pi(a_{t-1}|z_{t-1})\mathbb{P}(s_{t-1}|z_{t-1}) ds_{t-1} da_{t-1} ds_t}, \nonumber
        \end{align}
so $\mathbb{P}(s_t|z_t)$ depends on the policy $\pi$.

The posteriors $\mathbb{P}_{\pi}(z_t|s_t,a_t)$ and $\mathbb{P}_{\pi}(z_{t+1},z_t|s_{t+1},s_t)$ can be expressed using Bayes rule as
\begin{align}
    \mathbb{P}_{\pi}(z|s,a) &= \frac{\pi(a|z)\mathbb{P}(s|z)d_{\pi}(z)}{\rho_{\pi}(s,a)}, \nonumber \\
    \mathbb{P}_{\pi}(z,z'|s,s') &= \frac{\mathbb{P}(s,s'|z,z')\rho_{\pi}(z,z')}{\rho_{\pi}(s,s')}. \nonumber 
\end{align}
Without any additional assumptions, we cannot express these posteriors without an explicit dependence on the policy $\pi$ and, therefore, we cannot claim that they are policy independent.
\end{remark}

\begin{theorem}[From \citealt{rafailov2021visual}]
\label{theorem_3}
    Consider a POMDP and let $z_t$ be a latent state representation such that $\mathbb{P}(s_t|z_t, a_t) = \mathbb{P}(s_t|z_t) = \mathbb{P}(s_t|x_{\leq t}, a_{<t})$. Given $\mathbb{D}_f$ a generic $f$-divergence and under the assumption that the agent $\pi_{\bm{\theta}}$ and the expert $\pi_E$ share the same POMDP, the following inequality holds:
\begin{equation}
\mathbb{D}_f(\rho_{\pi_{\bm{\theta}}}(s,a), \rho_{\pi_E}(s,a)) \leq \mathbb{D}_f(\rho_{\pi_{\bm{\theta}}}(z,a), \rho_{\pi_E}(z,a)).
\end{equation}
\begin{proof}
We have that
\begin{align}
    \mathbb{D}_f(\rho_{\pi_{\bm{\theta}}}(s,a), \rho_{\pi_E}(s,a)) &= \mathbb{E}_{(s,a)\sim \rho_{\pi_E}(s,a)}\bigg[f\bigg(\frac{\rho_{\pi_{\bm{\theta}}}(s,a)}{\rho_{\pi_E}(s,a)}\bigg)\bigg] \nonumber \\
    &=\mathbb{E}_{(s,a)\sim \rho_{\pi_E}(s,a)}\bigg[f\bigg(\mathbb{E}_{z\sim \mathbb{P}_{\pi_{\bm{\theta}}}(z|s,a)}\bigg[\frac{\rho_{\pi_{\bm{\theta}}}(s,a)}{\rho_{\pi_E}(s,a)}\bigg]\bigg)\bigg] \nonumber \\
    &=\mathbb{E}_{(s,a)\sim \rho_{\pi_E}(s,a)}\bigg[f\bigg(\mathbb{E}_{z\sim\mathbb{P}_{\pi_{E}}(z|s,a)}\bigg[\frac{\rho_{\pi_{\bm{\theta}}}(s,a)}{\rho_{\pi_E}(s,a)} \frac{\mathbb{P}_{\pi_{\bm{\theta}}}(z|s,a)}{\mathbb{P}_{\pi_{E}}(z|s,a)}\bigg]\bigg)\bigg] \nonumber \\    
    &=\mathbb{E}_{(s,a)\sim \rho_{\pi_E}(s,a)}\bigg[f\bigg(\mathbb{E}_{z\sim\mathbb{P}_{\pi_{E}}(z|s,a)}\bigg[\frac{\rho_{\pi_{\bm{\theta}}}(s,a,z)}{\rho_{\pi_E}(s,a,z)}\bigg]\bigg)\bigg] \nonumber \\
    &\leq \mathbb{E}_{(s,a)\sim \rho_{\pi_E}(s,a)}\mathbb{E}_{z\sim\mathbb{P}_{\pi_{E}}(z|s,a)}\bigg[f\bigg(\frac{\rho_{\pi_{\bm{\theta}}}(s,a,z)}{\rho_{\pi_E}(s,a,z)}\bigg)\bigg] \label{theo_3_Jensen} \\
    &= \mathbb{E}_{(z,a)\sim \rho_{\pi_E}(z,a)}\mathbb{E}_{s\sim\mathbb{P}(s|z)}\bigg[f\bigg(\frac{\rho_{\pi_{\bm{\theta}}}(z,a)\mathbb{P}(s|z)}{\rho_{\pi_E}(z,a)\mathbb{P}(s|z)}\bigg)\bigg] \label{theo_3_P_s|z}\\
    &= \mathbb{D}_f(\rho_{\pi_{\bm{\theta}}}(z,a), \rho_{\pi_E}(z,a)). \nonumber
\end{align}

This proof follows by applying Jensen's inequality in \eqref{theo_3_Jensen} and by noticing that $\mathbb{P}(s|z)$ in \eqref{theo_3_P_s|z} does not depend on the policy but exclusively on the environment (Lemma~\ref{lemma_3}).
\end{proof}
\end{theorem}

\section{Theoretical analysis proofs}
\label{app:theo_proofs}

\paragraph{Theorem~\ref{theorem_1} restated:}
Consider a POMDP, and let $\mathcal{R}: \mathcal{S} \times \mathcal{A} \to \mathbb{R}$ and $z_t=\phi(x_{\leq t})$ such that $\mathbb{P}(s_t|z_t, a_t) = \mathbb{P}(s_t|z_t) = \mathbb{P}(s_t|x_{\leq t}, a_{<t})$. Then, the following inequality holds: 
\begin{align*}
    \begin{split}
        \big|J(\pi_{E}) - J(\pi_{\bm{\theta}})\big| \leq& \frac{2 R_{\max}}{1-\gamma}\mathbb{D}_{\text{\normalfont{TV}}}\big(\rho_{\pi_{\bm{\theta}}}(z,z'),\rho_{\pi_{E}}(z, z')\big) + C, 
    \end{split}
\end{align*}
where $R_{\max} = \max_{(s,a) \in \mathcal{S} \times \mathcal{A}}|\mathcal{R}(s,a)|$ and 
\begin{equation}
    C = \frac{2 R_{\max}}{1-\gamma}\mathbb{E}_{\rho_{\pi_{\bm{\theta}}}(z, z')}\big[\mathbb{D}_{\text{\normalfont{TV}}}\big(\mathbb{P}_{\pi_{\bm{\theta}}}(a|z, z'),\mathbb{P}_{\pi_{E}}(a|z, z')\big)\big].
    \label{app:theo_1_c}
\end{equation}

\begin{proof}
Note that $(1-\gamma)\cdot J(\pi) = \mathbb{E}_{(s,a)\sim\rho_{\pi}(s,a)}[\mathcal{R}(s,a)]$. Then,
\begin{align}
    \big|J(\pi_{E}) - J(\pi_{\bm{\theta}})\big| &= \frac{1}{1-\gamma}\bigg|\mathbb{E}_{(s,a)\sim\rho_{\pi_{E}}}[\mathcal{R}(s,a)] - \mathbb{E}_{(s,a)\sim\rho_{\pi_{\bm{\theta}}}}[\mathcal{R}(s,a)]\bigg| \nonumber \\ 
    &\leq \frac{1}{1-\gamma} \int_\mathcal{S}\int_\mathcal{A}\Big|\mathcal{R}(s,a)\Big|\Big|\rho_{\pi_E}(s, a) - \rho_{\pi_{\bm{\theta}}}(s, a)\Big|dads \nonumber \\
    &\leq \frac{R_{\max}}{1-\gamma} \int_\mathcal{S}\int_\mathcal{A}\Big|\rho_{\pi_E}(s, a) - \rho_{\pi_{\bm{\theta}}}(s, a)\Big|dads \nonumber \\
    &= \frac{2R_{\max}}{1-\gamma} \mathbb{D}_{\text{TV}}(\rho_{\pi_{\bm{\theta}}}(s,a),\rho_{\pi_{E}}(s,a)). \nonumber
\end{align}
It follows that
\begin{align}
    \big|J(\pi_{E}) - J(\pi_{\bm{\theta}})\big| &\leq \frac{2R_{\max}}{1-\gamma} \mathbb{D}_{\text{TV}}(\rho_{\pi_{\bm{\theta}}}(s,a),\rho_{\pi_{E}}(s,a)) \nonumber\\
    &\leq \frac{2R_{\max}}{1-\gamma} \mathbb{D}_{\text{TV}}(\rho_{\pi_{\bm{\theta}}}(z,a),\rho_{\pi_{E}}(z,a)) \label{proof:step_2}\\
    &\leq \frac{2R_{\max}}{1-\gamma}\mathbb{D}_{\text{TV}}\big(\rho_{\pi_{\bm{\theta}}}(z, z'),\rho_{\pi_E}(z, z')\big) + C, \label{proof:step_5}
\end{align}
where $C$ is in \eqref{app:theo_1_c}.
Note that \eqref{proof:step_2} follows from Theorem~\ref{theorem_3} and \eqref{proof:step_5} from Lemma~\ref{lemma_2}. 
\end{proof}

\begin{corollary}
\label{corollary}
    Consider the same conditions as in Theorem~\ref{theorem_1}. Provided the injectivity of $\mathbb{P}(z'|z,a)$ with respect to $a$, then 
\begin{align*}
    \begin{split}
    \big|J(\pi_{E}) - J(\pi_{\bm{\theta}})\big| 
    \leq \frac{2 R_{\max}}{1-\gamma}\mathbb{D}_{\text{\normalfont{TV}}}\big(\rho_{\pi_{\bm{\theta}}}(z, z'),\rho_{\pi_{E}}(z, z')\big).
\end{split}
\end{align*}
\begin{proof}
    The proof follows similarly to \citet{yang2019imitation} in the MDP case. Recall 
\begin{equation*}
    \mathbb{P}_{\pi_{\bm{\theta}}}(a|z, z') = \frac{\mathbb{P}(z'|z, a)\pi_{\bm{\theta}}(a|z)}{\int_{\mathcal{A}} \mathbb{P}(z'|z,\Bar{a})\pi_{\bm{\theta}}(\Bar{a}|z)d\Bar{a}},
\end{equation*} 
and 
\begin{align}
    \begin{split}
    \mathbb{P}(z'|z,a)=& \int_{\mathcal{S}}\int_{\mathcal{S}}\int_{\mathcal{X}}\mathbb{P}(z'|x',a,z)\mathcal{U}(x'|s')\mathcal{T}(s'|s,a)\mathbb{P}(s|z) dx' ds' ds. \nonumber
    \end{split}
\end{align}
We can write $\mathbb{P}(z'|z, a)=\delta(z' - f(z,a))$, where $f:\mathcal{Z} \times \mathcal{A} \to \mathcal{Z}$ and $\delta$ is a Dirac delta function. It follows that 
\begin{align*}
    \mathbb{P}_{\pi_{\bm{\theta}}}(a|z, z') = \mathbb{P}_{\pi_E}(a| z, z') = \delta(z' - f(z,a)),
\end{align*}
which yields
\begin{align*}
\mathbb{E}_{(z,z')\sim\rho_{\pi_{\bm{\theta}}}(z,z')}\bigg[\mathbb{D}_{\text{TV}}\big(\mathbb{P}_{\pi_{\bm{\theta}}}(a| z, z'),\mathbb{P}_{\pi_E}(a| z, z')\big)\bigg] = 0.
\end{align*}
Using this result in Theorem~\ref{theorem_1} proves the corollary.
\end{proof}
\end{corollary}

\paragraph{Theorem~\ref{theorem_2} restated:}
Consider a POMDP, and let $\mathcal{R}: \mathcal{S} \times \mathcal{S} \to \mathbb{R}$ and $z_t=\phi(x_{\leq t})$ such that $\mathbb{P}(s_t|z_t, a_t) = \mathbb{P}(s_t|z_t) = \mathbb{P}(s_t|x_{\leq t}, a_{<t})$. Then, the following inequality holds:
\begin{align*}
    \begin{split}
        \big|J(\pi_{E}) - J(\pi_{\bm{\theta}})\big| \leq& \frac{2 R_{\max}}{1-\gamma}\mathbb{D}_{\text{\normalfont{TV}}}\big(\rho_{\pi_{\bm{\theta}}}(z,z'),\rho_{\pi_{E}}(z, z')\big),
    \end{split}
\end{align*}
where $R_{\max} = \max_{(s,s') \in \mathcal{S} \times \mathcal{S}}|\mathcal{R}(s,s')|$.
\begin{proof}
Note that $(1-\gamma)\cdot J(\pi) = \mathbb{E}_{(s,s')\sim\rho_{\pi}(s,s')}[\mathcal{R}(s,s')]$. Then,
\begin{align}
    \big|J(\pi_{E}) - J(\pi_{\bm{\theta}})\big| &= \frac{1}{1-\gamma}\bigg|\mathbb{E}_{(s,s')\sim\rho_{\pi_{E}}}[\mathcal{R}(s,s')] - \mathbb{E}_{(s,s')\sim\rho_{\pi_{\bm{\theta}}}}[\mathcal{R}(s,s')]\bigg| \nonumber \\ 
    &\leq \frac{1}{1-\gamma} \int_\mathcal{S}\int_\mathcal{S}\Big|\mathcal{R}(s,s')\Big|\Big|\rho_{\pi_E}(s, s') - \rho_{\pi_{\bm{\theta}}}(s, s')\Big|ds'ds \nonumber \\
    &\leq \frac{R_{\max}}{1-\gamma} \int_\mathcal{S}\int_\mathcal{S}\Big|\rho_{\pi_E}(s, s') - \rho_{\pi_{\bm{\theta}}}(s, s')\Big|ds'ds \nonumber \\
    &= \frac{2 R_{\max}}{1-\gamma} \mathbb{D}_{\text{TV}}(\rho_{\pi_{\bm{\theta}}}(s,s'),\rho_{\pi_{E}}(s,s')) \nonumber.
\end{align}
In order to conclude the proof, we have to show that 
\begin{equation}
    \mathbb{D}_{\text{TV}}\big(\rho_{\pi_{\bm{\theta}}}(s,s'),\rho_{\pi_{E}}(s,s')\big) \leq \mathbb{D}_{\text{TV}}\big(\rho_{\pi_{\bm{\theta}}}(z,z'),\rho_{\pi_{E}}(z,z')\big).
    \label{theo_1_KL}
\end{equation}
Consider a generic $f$-divergence as in Definition~\ref{def:f_div}. Then,
\begin{align}
    &\mathbb{D}_f(\rho_{\pi_{\bm{\theta}}}(s,s'), \rho_{\pi_E}(s,s')) \nonumber \\
    &\quad= \mathbb{E}_{(s,s')\sim \rho_{\pi_E}(s,s')}\bigg[f\bigg(\frac{\rho_{\pi_{\bm{\theta}}}(s,s')}{\rho_{\pi_E}(s,s')}\bigg)\bigg] \nonumber \\
    &\quad=\mathbb{E}_{(s,s')\sim \rho_{\pi_E}(s,s')}\bigg[f\bigg(\mathbb{E}_{(z,z')\sim\mathbb{P}_{\pi_{\bm{\theta}}}(z,z'|s,s')}\bigg[\frac{\rho_{\pi_{\bm{\theta}}}(s,s')}{\rho_{\pi_E}(s,s')}\bigg]\bigg)\bigg] \nonumber \\
    &\quad=\mathbb{E}_{(s,s')\sim \rho_{\pi_E}(s,s')}\bigg[f\bigg(\mathbb{E}_{(z,z')\sim\mathbb{P}_{\pi_{E}}(z,z'|s,s')}\bigg[\frac{\rho_{\pi_{\bm{\theta}}}(s,s')}{\rho_{\pi_E}(s,s')}\frac{\mathbb{P}_{\pi_{\bm{\theta}}}(z,z'|s,s')}{\mathbb{P}_{\pi_{E}}(z,z'|s,s')}\bigg]\bigg)\bigg] \nonumber \\    
    &\quad=\mathbb{E}_{(s,s')\sim \rho_{\pi_E}(s,s')}\bigg[f\bigg(\mathbb{E}_{(z,z')\sim\mathbb{P}_{\pi_{E}}(z,z'|s,s')}\bigg[\frac{\rho_{\pi_{\bm{\theta}}}(s,s',z,z')}{\rho_{\pi_E}(s,s',z,z')}\bigg]\bigg)\bigg] \nonumber \\
    &\quad\leq \mathbb{E}_{(s,s')\sim \rho_{\pi_E}(s,s')}\mathbb{E}_{(z,z')\sim\mathbb{P}_{\pi_{E}}(z,z'|s,s')}\bigg[f\bigg(\frac{\rho_{\pi_{\bm{\theta}}}(s,s',z,z')}{\rho_{\pi_E}(s,s',z,z')}\bigg)\bigg] \label{theo_1_Jensen} \\
    &\quad= \mathbb{E}_{(z,z')\sim \rho_{\pi_E}(z,z')}\mathbb{E}_{(s,s')\sim\mathbb{P}(s,s'|z,z')}\bigg[f\bigg(\frac{\rho_{\pi_{\bm{\theta}}}(z,z')\mathbb{P}(s,s'|z,z')}{\rho_{\pi_E}(z,z')\mathbb{P}(s,s'|z,z')}\bigg)\bigg] \label{theo_1_P_s|z} \nonumber \\
    &\quad= \mathbb{D}_f(\rho_{\pi_{\bm{\theta}}}(z,z'), \rho_{\pi_E}(z,z')), 
\end{align}
from which \eqref{theo_1_KL} follows by considering $f(x)=\frac{1}{2}|x-1|$. Note that \eqref{theo_1_Jensen} uses Jensen's inequality and \eqref{theo_1_P_s|z} follows considering that $\mathbb{P}(s,s'|z,z')$ is policy independent (Lemma~\ref{lemma_3}). Hence, $\mathbb{P}(s,s'|z,z')$ depends only on the environment which is, by assumption, the same for both the expert and the agent.
\end{proof}

\newpage
\section{LAIfO algorithm and hyperparameters}
\label{app:LAIfO}

\begin{algorithm}[H]
\label{alg:LAIfO}
\caption{LAIfO}
\textbf{Inputs}: \\
Expert observations: $(x_n)_{0:N} \in \mathcal{B}_{E}$. \\
$\pi_{\bm{\theta}}$, $D_{\bm{\chi}}$, $Q_{\bm{\psi}_1}$, $Q_{\bm{\psi}_2}$, $\phi_{\bm{\delta}}$: networks for policy, discriminator, Q functions, and feature extractor. \\
$T_{\text{train}}$, $\sigma(t)$, $d$, aug, $c$, $\tau$, $B$, $\alpha$, $\alpha_D$, $\lambda$, $\gamma$: training steps, scheduled standard deviation, frames stack dimension, data augmentation,  clip value, target update rate, batch size, learning rate, discriminator learning rate, discriminator gradient penalty weight, and discount factor. \\
\For{$t=1, \dots, T_{\text{\normalfont{train}}}$}{
$\sigma_t \leftarrow \sigma(t)$ \\
\If{$t \geq d-1$}{$z_t \leftarrow \phi_{\bm{\delta}}(x_{t-d+1:t})$}
\Else{$z_t \leftarrow \phi_{\bm{\delta}}(x_{0:t})$} 
$a_t \leftarrow \pi_{\bm{\theta}}(z_t) + \epsilon$ and $\epsilon \sim \mathcal{N}(0,\sigma_t^2)$ \\
$s_{t+1} \sim \mathcal{T}(\cdot|s_t, a_t)$ and $x_{t+1} \sim \mathcal{U}(\cdot|s_{t+1})$ \\
$\mathcal{B} \leftarrow \mathcal{B} \cup (x_t, a_t, x_{t+1})$ \\
UpdateDiscriminator($\mathcal{B}$, $\mathcal{B}_E$) \\
UpdateCritic($\mathcal{B}$) \\
UpdateActor($\mathcal{B}$)\\
}
\Begin(UpdateDiscriminator){
$\{(x_{t-d+1:t}, x_{t-d+2:t+1})\} \sim \mathcal{B}_{E}$ and $\{(x_{t-d+1:t}, x_{t-d+2:t+1})\} \sim \mathcal{B}$  (sample $B$ transitions)\\
$z_t \leftarrow \phi_{\bm{\delta}}(\text{aug}(x_{t-d+1:t}))$ and $z_{t+1} \leftarrow \phi_{\bm{\delta}}(\text{aug}(x_{t-d+2:t+1}))$ for both agent and expert \\
Update $D_{\bm{\chi}}$ to minimize \eqref{eq:AIL_BCE} with learning rate $\alpha_D$ and gradient penalty weight $\lambda$
}
\Begin(UpdateCritic){
$\{(x_{t-d+1:t}, a_t, x_{t-d+2:t+1})\} \sim \mathcal{B}$ (sample $B$ transitions)\\ 
$z_t \leftarrow \phi_{\bm{\delta}}(\text{aug}(x_{t-d+1:t}))$ and $z_{t+1} \leftarrow \phi_{\bm{\delta}}(\text{aug}(x_{t-d+2:t+1}))$ \\
$a_{t+1} \leftarrow $ $\pi_{\bm{\theta}}(z_{t+1}) + \epsilon$ and $\epsilon \sim \text{clip}(\mathcal{N}(0,\sigma_t^2), -c, c)$ \\
Update $Q_{\bm{\psi}_1}$, $Q_{\bm{\psi}_2}$ and $\phi_{\bm{\delta}}$ to minimize \eqref{eq:DDPG_critic} with $r_{\bm{\chi}}(z_t, z_{t+1})$ in \eqref{eq:AIL_BCE_reward} and learning rate $\alpha$ \\
$\bar{\bm{\psi}}_k \leftarrow (1 - \tau)\bar{\bm{\psi}}_k +\tau\bm{\psi}_k \ \ \ \forall k \in \{1,2\}$
}
\Begin(UpdateActor){
$\{x_{t-d+1:t}\} \sim \mathcal{B}$ (sample $B$ observations)\\
$z_t \leftarrow \phi_{\bm{\delta}}(\text{aug}(x_{t-d+1:t}))$ \\ 
$a_t \leftarrow \pi_{\bm{\theta}}(z_t) + \epsilon$ and $\epsilon \sim \text{clip}(\mathcal{N}(0,\sigma_t^2), -c, c)$ \\
Update $\pi_{\bm{\theta}}$ using DDPG~\citep{lillicrap2015continuous} with learning rate $\alpha$
}
\end{algorithm}

\begin{table}[h!]
\centering
\caption{Hyperparameter values for LAIfO experiments.}
\label{tab:Hyper_1}
\small
\begin{tabular}{c c c c}\toprule
\multicolumn{2}{l}{Hyperparameter Name} & \multicolumn{2}{c}{Value}\\
\cmidrule(lr){1-2} \cmidrule(lr){3-4}
\multicolumn{2}{l}{Frames stack $(d)$} & \multicolumn{2}{c}{$3$} \\
\multicolumn{2}{l}{Discount factor $(\gamma)$} & \multicolumn{2}{c}{$0.99$} \\
\multicolumn{2}{l}{Image size} & \multicolumn{2}{c}{$84 \times 84$} \\
\multicolumn{2}{l}{Batch size $(B)$} & \multicolumn{2}{c}{$256$} \\
\multicolumn{2}{l}{Optimizer} & \multicolumn{2}{c}{Adam}\\
\multicolumn{2}{l}{Augmentation} & \multicolumn{2}{c}{Crop}\\
\multicolumn{2}{l}{Learning rate $(\alpha)$} & \multicolumn{2}{c}{$10^{-4}$}\\
\multicolumn{2}{l}{Discriminator learning rate $(\alpha_D)$} & \multicolumn{2}{c}{$4 \times 10^{-4}$}\\
\multicolumn{2}{l}{Discriminator gradient penalty weight $(\lambda)$} & \multicolumn{2}{c}{$10$}\\
\multicolumn{2}{l}{Target update rate $(\tau)$} & \multicolumn{2}{c}{$0.01$}\\
\multicolumn{2}{l}{Clip value $(c)$} & \multicolumn{2}{c}{$0.3$}\\
\bottomrule
\end{tabular}
\end{table}

\newpage
\section{Experiments curves and additional ablations}
\label{app:learning_curves}

All the experiments are run using Nvidia-A40 GPUs on an internal cluster. For each algorithm, we run two experiments in parallel on the same GPU and each experiment takes 1 to 10 days depending on the simulated environment and the considered algorithm. For all the implementation details refer to our \href{https://github.com/VittorioGiammarino/AIL_from_visual_obs/tree/LAIfO}{code}\footnote{https://github.com/VittorioGiammarino/AIL\_from\_visual\_obs/tree/LAIfO}. 

\begin{figure}[t]
    \centering
    \includegraphics[width=0.9\linewidth]{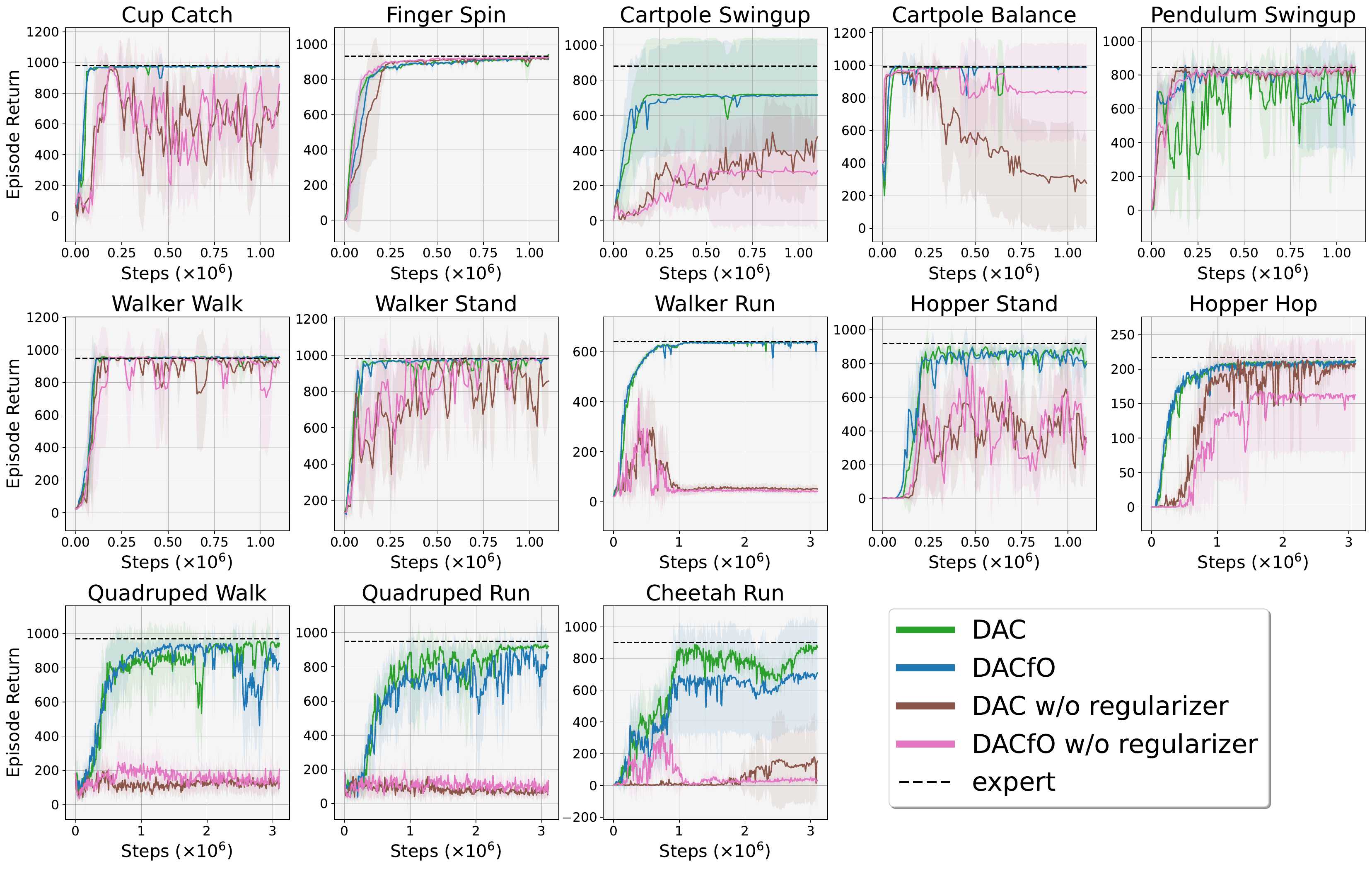}
    \caption{Learning curves for the fully observable IL and IfO settings in Table~\ref{table_IL-IfO}. Plots show the average return per episode as a function of training steps.}
    \label{fig:MDP_app}
\end{figure}

\begin{figure}
    \centering
    \includegraphics[width=0.9\linewidth]{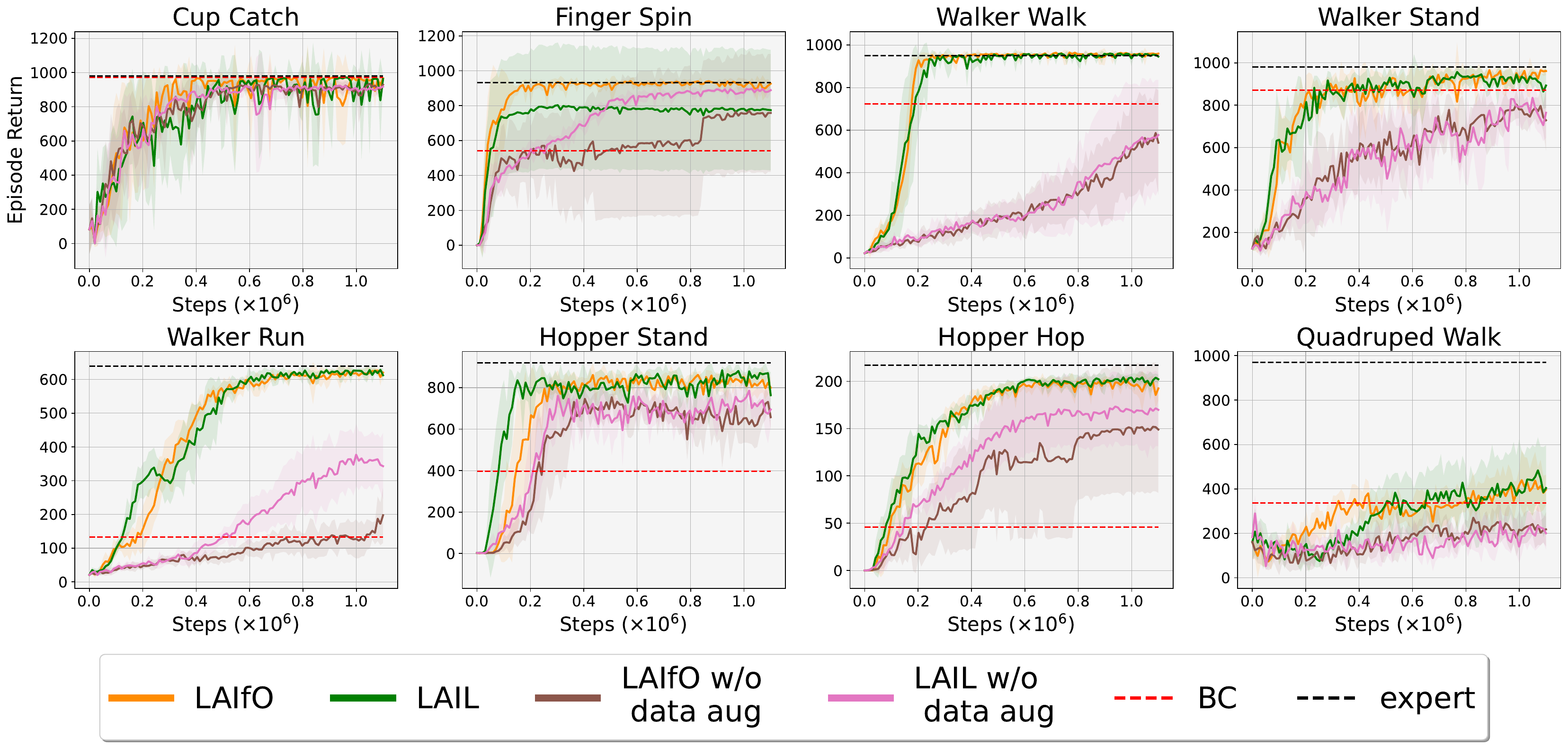}
    \caption{Learning curves for the results in Table~\ref{table_data_augmentation_ablation}. Plots show the average return per episode as a function of training steps.}
    \label{fig:data_aug_ablation}
\end{figure}

\begin{figure}
    \centering
    \includegraphics[width=0.9\linewidth]{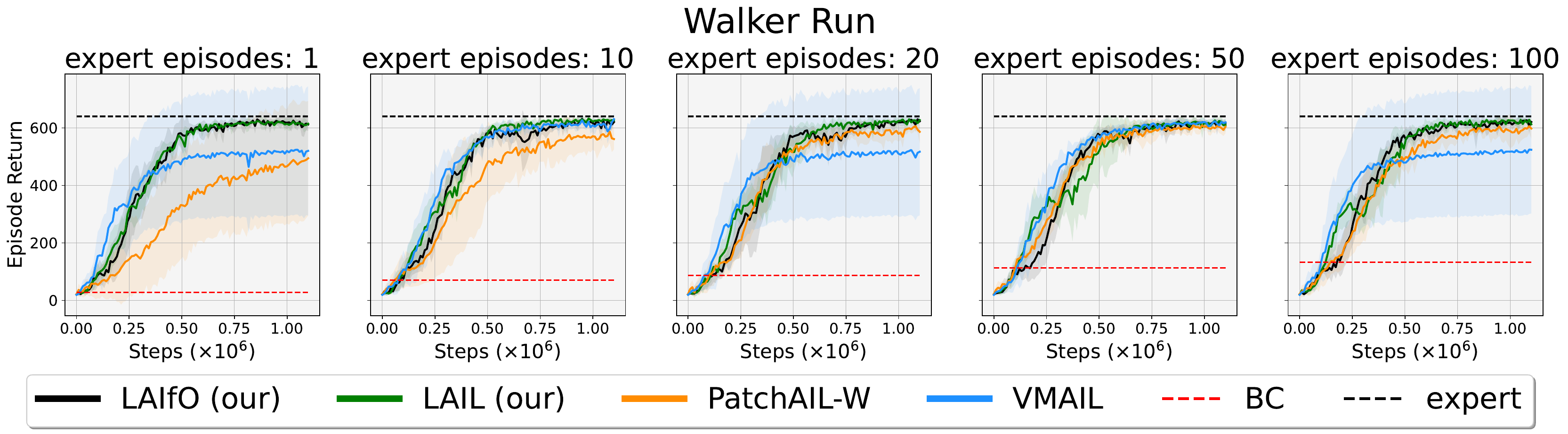}
    \caption{Learning curves for the results in Table~\ref{table_n_expert}. Plots show the average return per episode as a function of training steps.}
    \label{fig:n_experts_episodes}
\end{figure}

Table~\ref{table_IL-IfO} summarizes the asymptotic performance of DAC and DACfO in the fully observable setting. We highlight in Table~\ref{table_IL-IfO} the importance of the discriminator penalty term in \eqref{eq:penalty_grad_reg} by comparing DAC and DACfO to results that do not include this regularizer in the adversarial loss in \eqref{eq:AIL_BCE}. Figure~\ref{fig:MDP_app} shows the full learning curves as a function of training steps for Table~\ref{table_IL-IfO}. The results of DAC and DACfO in Table~\ref{table_IL-IfO} are used in Figure~\ref{fig:cum_hist}. Moreover, Table~\ref{table_data_augmentation_ablation} and Figure~\ref{fig:data_aug_ablation} show the results for the ablation study on data augmentation in LAIL and LAIfO. These results highlight the importance of data augmentation on the latent variable $z \in \mathcal{Z}$ inference problem, as we observe a decrease in performance when data augmentation is not used. 
Finally, Figure~\ref{fig:n_experts_episodes} shows the full learning curves for the results in Table~\ref{table_n_expert}.


\begin{table}
\centering
\small
\caption{Experimental results for IL and IfO (i.e., imitation in fully observable environments, with and without access to expert actions, respectively). We use DDPG to train experts in a fully observable setting and collect $100$ episodes of expert data. The experiments are conducted as in Table~\ref{table_V-IfO}. We report mean and standard deviation of final performance over seeds.}
\label{table_IL-IfO}
    \begin{tabular}{c | c | c c c c }\toprule
        & Expert & \makecell{DAC \\ \citep{kostrikov2018discriminator}} & DACfO & \makecell{DAC w/o \\ regularizer in \eqref{eq:penalty_grad_reg}} & \makecell{DACfO w/o \\ regularizer in \eqref{eq:penalty_grad_reg}} \\ 
        \cmidrule(lr){1-6}
        Cup Catch & $980$ & \bm{$974 \pm 2.2$} & $971 \pm 3.9$ & $746 \pm 302$ & $859 \pm 143$ \\
        Finger Spin & $932$ & \bm{$939 \pm 14.2$} & $913 \pm 11.8$ & $918 \pm 3.1$ & $925 \pm 4.9$ \\
        Cartpole Swingup & $881$ & \bm{$717 \pm 321$} & $715 \pm 320$ & $477 \pm 259$ & $284 \pm 312$ \\
        Cartpole Balance & $990$ & \bm{$989 \pm 2.3$} & $987 \pm 1.1$ & $278 \pm 265$ & $837 \pm 303$ \\
        Pendulum Swingup & $845$ & $849 \pm 25.1$ & $621 \pm 320$ & $832 \pm 38.5$ & \bm{$851 \pm 26.0$} \\
        Walker Walk & $960$ & \bm{$957 \pm 5.1$} & \bm{$957 \pm 4.3$} & $920 \pm 40.1$ & $919 \pm 67.8$ \\
        Walker Stand & $980$ &  $985 \pm 3.0$ & $982 \pm 4.6$ & $857 \pm 192$ & \bm{$986 \pm 3.0$} \\
        Walker Run & $640$ &  \bm{$636 \pm 4.4$} & \bm{$636 \pm 4.4$} & $51.5 \pm 15.5$ & $42.2 \pm 9.8$ \\
        Hopper Stand & $920$ & \bm{$812 \pm 126$} & $796 \pm 176$ & $335 \pm 152$ & $361 \pm 138$ \\
        Hopper Hop & $217$ & \bm{$213 \pm 2.0$} & $212 \pm 1.9$ & $208 \pm 6.7$ & $163 \pm 81.8$ \\
        Quadruped Walk & $970$ & \bm{$943 \pm 29.0$} & $826 \pm 175$ & $126 \pm 50.1$ & $207 \pm 73.8$ \\
        Quadruped Run & $950$ & \bm{$919 \pm 13.4$} & $870 \pm 91.7$ & $109 \pm 62.3$ & $133 \pm 61.6$ \\
        Cheetah Run & $900$ & \bm{$871 \pm 33.9$} & $710 \pm 355$ & $152 \pm 288$ & $31.4 \pm 20.4$ \\
        \bottomrule
    \end{tabular}
\end{table}

\begin{table}
\centering
\small
\caption{Ablation study on the importance of data augmentation in LAIfO and LAIL. We use DDPG to train experts in a fully observable setting and collect $100$ episodes of expert data. The experiments are conducted as in Table~\ref{table_V-IfO}. We report mean and standard deviation of final performance over seeds after training for $10^6$ frames.}
\label{table_data_augmentation_ablation}
    \begin{tabular}{c | c  c | c c c c }\toprule
        & Expert & BC & \makecell{LAIfO} & LAIL & \makecell{LAIfO w/o \\ data augmentation} & \makecell{LAIL w/o \\ data augmentation} \\ 
        \cmidrule(lr){1-7}
        Cup Catch & $980$ & $971 \pm 36.0$ & \bm{$967 \pm 7.6$} & $962 \pm 18.0$ & $929 \pm 35.8$ & $918 \pm 36.0$ \\
        Finger Spin & $932$ & $542\pm 219 $ & \bm{$926 \pm 10.7$} & $775 \pm 345$ & $758 \pm 339$ & $889 \pm 12.8$ \\
        Walker Walk & $960$ & $723 \pm 137$ & \bm{$960 \pm 2.2$} & $946 \pm 8.6$ & $541 \pm 247$ & $577 \pm 259$ \\
        Walker Stand & $960$ & $871 \pm 77.7$ & \bm{$961 \pm 20.0$} & $893 \pm 106$ & $729 \pm 144$ & $765 \pm 118$ \\
        Walker Run & $640$ & $133 \pm 27.8$ & $618 \pm 4.6$ & \bm{$625 \pm 5.1$} & $198 \pm 82.4$ & $327 \pm 100$ \\
        Hopper Stand & $920$ & $398 \pm 96.4$ & \bm{$800 \pm 46.7$} & $764 \pm 111$ & $658 \pm 60.8$ & $695\pm 81.0$ \\
        Hopper Hop & $217$ & $45.9 \pm 22.1$ & $193 \pm 12.9$ & \bm{$202 \pm 4.8$} & $149 \pm 66.9$ & $170 \pm 38.5$ \\
        Quadruped Walk & $970$ & $337 \pm 50.5$ & $397 \pm 67.0$ & \bm{$404 \pm 192$} & $218 \pm 77.2$ & $201 \pm 73.0$ \\
        \bottomrule
    \end{tabular}
\end{table}

\end{document}